\documentclass{article} 

\PassOptionsToPackage{table}{xcolor}

\usepackage{iclr2026_conference,times}

\usepackage{amsmath,amsfonts,bm}

\def\eqref#1{equation~\ref{#1}}

\def\1{\bm{1}}

\DeclareMathAlphabet{\mathsfit}{\encodingdefault}{\sfdefault}{m}{sl}
\SetMathAlphabet{\mathsfit}{bold}{\encodingdefault}{\sfdefault}{bx}{n}

\usepackage{url}
\usepackage{multirow}
\usepackage{pifont}
\usepackage{booktabs}
\usepackage{makecell}     
\usepackage{array}
\usepackage{amssymb}
\usepackage{graphicx}
\usepackage{wrapfig}
\definecolor{lightblue}{RGB}{100,140,210}
\definecolor{myblue}{RGB}{33, 84, 151}
\definecolor{lightgreen}{RGB}{199,232,181}
\definecolor{mypurple}{RGB}{148,0,211}
\definecolor{varysorange}{RGB}{255,140,0}
\usepackage{caption}

\usepackage[ruled,vlined]{algorithm2e}

\usepackage{amsmath,amsthm} 
\newtheorem{proposition}{Proposition}
\newtheorem{theorem}{Theorem}
\newtheorem{corollary}{Corollary}

\title{SWIRL: A Staged Workflow for Interleaved Reinforcement Learning in Mobile GUI Control}

\author{Quanfeng Lu$^{1}$\footnotemark[1] , Zhantao Ma$^{1}$\thanks{Equal Contribution} , Shuai Zhong$^{1}$ , Jin Wang$^{1}$ , Dahai Yu$^{3}$ , Michael K. Ng$^{2}$ , Ping Luo$^{1}$\thanks{Corresponding Author: pluo@cs.hku.hk} \\
$^{1}$The University of Hong Kong \quad$^{2}$Hong Kong Baptist University \\
$^{3}$TCL Corporate Research (Hong Kong) Co., Ltd\\
\url{https://github.com/Lqf-HFNJU/SWIRL}
}

\iclrfinalcopy 
\begin{document}

\maketitle

\begin{abstract}

The rapid advancement of large vision language models (LVLMs) and agent systems has heightened interest in mobile GUI agents that can reliably translate natural language into interface operations. Existing single-agent approaches, however, remain limited by structural constraints. Although multi-agent systems naturally decouple different competencies, recent progress in multi-agent reinforcement learning (MARL) has often been hindered by inefficiency and remains incompatible with current LVLM architectures. To address these challenges, we introduce SWIRL, a staged workflow for interleaved reinforcement learning designed for multi-agent systems. SWIRL reformulates MARL into a sequence of single-agent reinforcement learning tasks, updating one agent at a time while keeping the others fixed. This formulation enables stable training and promotes efficient coordination across agents. Theoretically, we provide a stepwise safety bound, a cross-round monotonic improvement theorem, and convergence guarantees on return, ensuring robust and principled optimization. In application to mobile GUI control, SWIRL instantiates a Navigator that converts language and screen context into structured plans, and an Interactor that grounds these plans into executable atomic actions. Extensive experiments demonstrate superior performance on both high-level and low-level GUI benchmarks. Beyond GUI tasks, SWIRL also demonstrates strong capability in multi-agent mathematical reasoning, underscoring its potential as a general framework for developing efficient and robust multi-agent systems.

\end{abstract}

\section{Introduction}


With the rapid progress of large vision–language models (LVLMs)~\citep{openai2025introducinggpt5, zhu2025internvl3, Qwen2.5-VL, guo2025seed1}, increasing attention has been devoted to mobile GUI agents capable of translating natural language instructions into reliable interface manipulation~\citep{qin2025ui, xu2024aguvis, wu2024atlas, lu2024gui, luo2025gui}. These agents ground user instructions in the current screenshot and interaction history, reason over this evolving state, and iteratively generate the next action until the task is completed. Effective mobile GUI control depends on two key competencies: task planning, which forms global, goal-conditioned decisions under evolving contexts, and task execution, which translates these plans into executable actions with precise localization. Most existing systems adopt a single-agent design, which complicates the robust integration of these competencies.


We identify two fundamental challenges. First, coupling high-level planning with fine-grained perception and precise actuation makes single end-to-end policies prone to interference and brittleness~\citep{wang2024gui, erdogan2025plan, mo2025building, wang2025mp}. Second, end-to-end systems often exhibit a weak linkage between reasoning traces and executed actions, sometimes producing correct outcomes for spurious reasons or plausible traces paired with faulty actions~\citep{turpin2023language, arcuschin2025chain, li2024towards}, thereby undermining safety and accountability in assurance-critical applications~\citep{zhang2025does, shi2025towards, kuntz2025harm}.
\begin{figure*}[!t]
    \centering
    \includegraphics[width=0.85\textwidth]{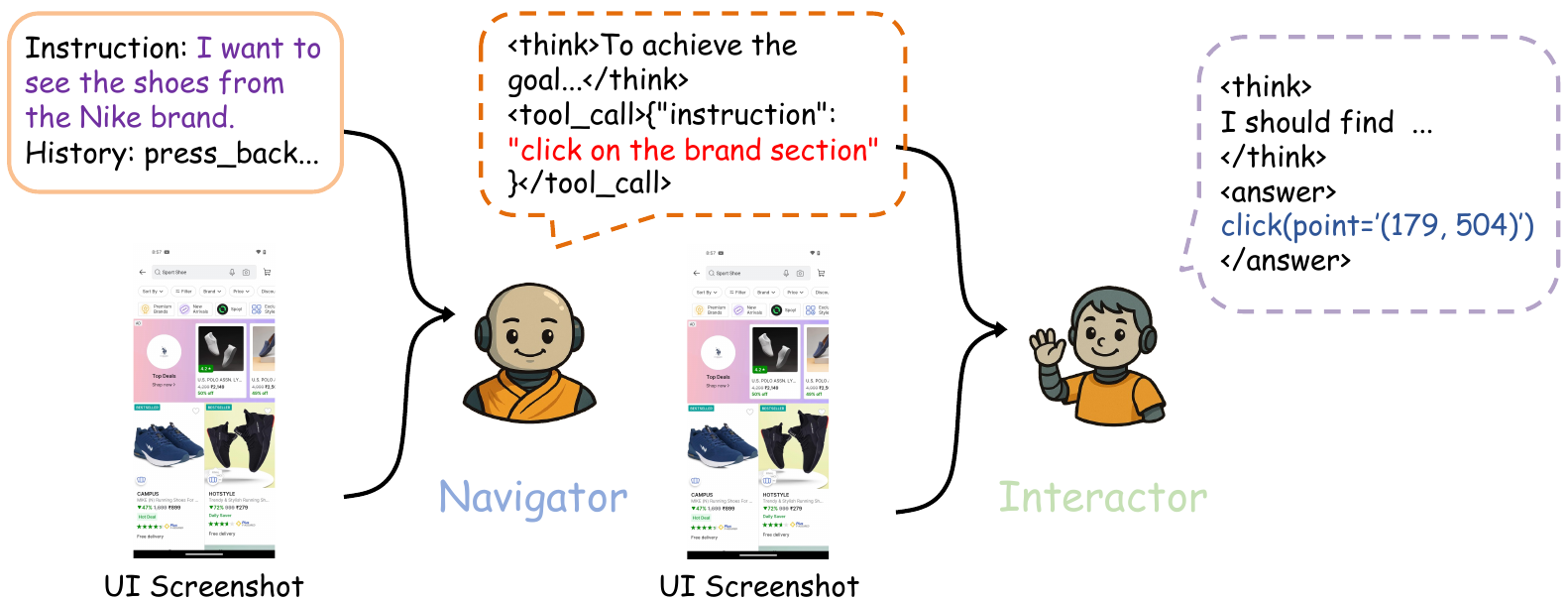}
    \caption{Our multi-agent inference pipelines. Given a \textcolor{mypurple}{high-level instruction}, UI screenshots, and historical actions, the Navigator generates a \textcolor{red}{low-level instruction}, which the Interactor then uses together with UI screenshots to produce the final \textcolor{myblue}{executable action}. This design decouples task planning from execution, enabling specialization, and leverages explicit intermediate instructions to enhance transparency and interpretability.}
    \label{fig:varys_infer}
    \vspace{-2mm}
\end{figure*}


A multi-agent design provides a principled approach to decoupling core competencies by assigning planning and execution to specialized agents. 
Beyond this division of labor, structured interactions between agents further enhance transparency by making the reasoning process more interpretable and the resulting actions more auditable. This explicit linkage between decision-making and execution not only improves accountability but also facilitates supervision and error analysis, which are essential for building reliable GUI control systems.
However, training-free adaptation of generic LVLMs rarely suffices for domain-specific GUI control due to insufficient cooperation~\citep{niu2024screenagent}. Naive multi-agent reinforcement learning (MARL) further introduces practical challenges: joint optimization of multiple policies inflates compute budgets and limited capabilities~\citep{wang2022model,gogineni2023towards}. Meanwhile, high-throughput reinforcement learning (RL) frameworks developed for LVLMs are almost exclusively engineered for single-agent training~\citep{hu2024openrlhf, sheng2025hybridflow}, making them ill-suited for MARL~\citep{liao2025marft}. These limitations motivate a central question: \textit{can we train multi-agent systems that are resource-friendly while ensuring provable stability during training?}
\begin{wrapfigure}{r}{0.5\textwidth}
    \centering
    \includegraphics[width=0.5\textwidth]{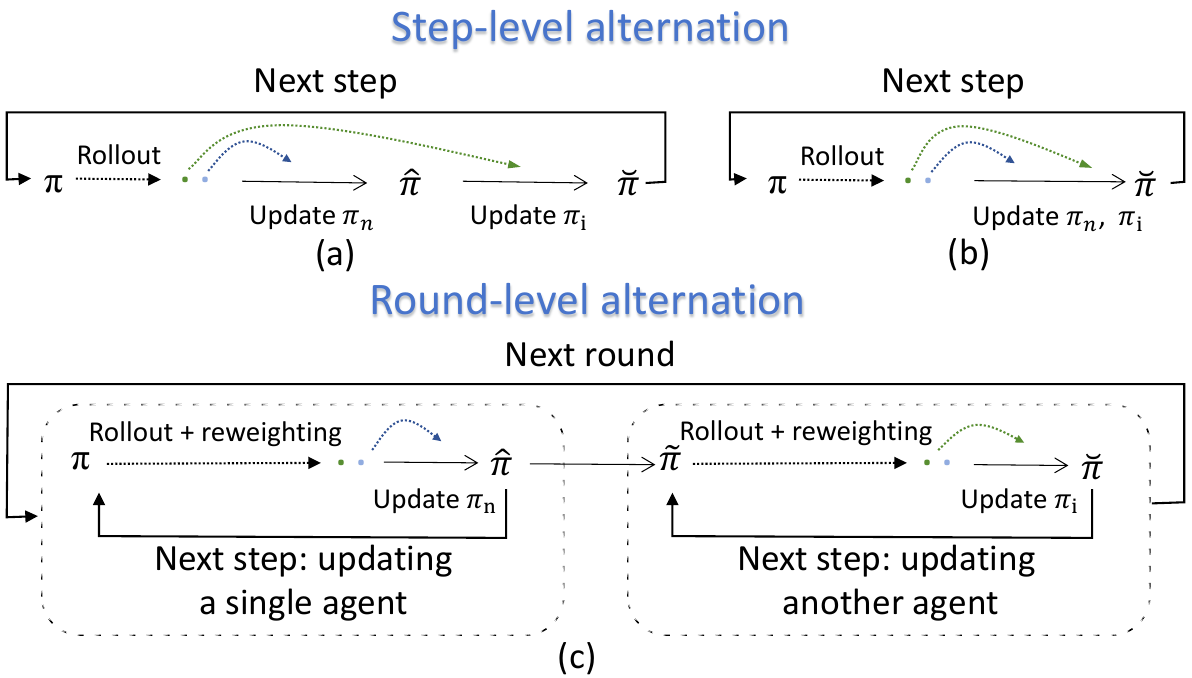}
    \caption{Alternating training paradigm. (a) HAPPO~\citep{zhong2024heterogeneous}: step-level single-agent sequential updates; (b) A2PO~\citep{wang2023order} \& MARFT~\citep{liao2025marft}: enhancing sequential updates with preceding-agent off-policy correction for greater efficiency; (c) SWIRL: round-level alternation with inner solves.}
    \label{fig:compare of SWIRL}
\end{wrapfigure}

We introduce SWIRL, a staged workflow for interleaved reinforcement learning. SWIRL decomposes multi-agent training into two phases: independent pre-warming of each module, followed by alternating optimization where one module is frozen while the other is updated. During this alternating process, we further incorporate an online reweighting mechanism to enhance training stability and accelerate convergence. SWIRL in Fig.~\ref{fig:compare of SWIRL}(c) updates exactly one agent at a step. After several updates on one agent, it then switches to the next, turning joint optimization into sequential single-agent problems and enabling reuse of standard RL toolchains in agent training while maintaining effective coordination. 
Beyond practicality, we offer theoretical and system-level benefits: we establish a per-step safety bound, prove monotonic improvement across rounds, and derive a corollary for return convergence. In implementation, SWIRL requires only the currently updated agent to be resident on the training device, yielding \(O(1)\) actor memory usage, smooth compatibility with standard stacks, and support for heterogeneous model sizes and update budgets, contrasting with the \(O(N)\) memory usage of other methods. 
Table~\ref{tab:params} details the count of actor parameters loaded during training without extra optimizations.

We instantiate SWIRL on mobile GUI control with a dual-agent architecture. 
The Navigator interprets instructions, interaction history, and the current screenshot to form a task context and produce structured low-level instructions (LLI). The Interactor consumes the LLI together with the current UI view and outputs atomic actions such as click, scroll, and text input with precise localization. 
\begin{wraptable}{r}{0.35\linewidth} 
  \centering
  \caption{Comparison of actor parameters loaded during training.}
  \label{tab:params}
  \scalebox{0.8}{%
    \begin{tabular}{lc}
      \toprule
      \textbf{Method} & \textbf{Actor parameters} \\
      \midrule
      HAPPO & $\sum_{i=1}^{N}\lvert\theta_{i}\rvert$ \\
      A2PO  & $\sum_{i=1}^{N}\lvert\theta_{i}\rvert$ \\
      MARFT        & $\sum_{i=1}^{N}\lvert\theta_{i}\rvert$ \\
      \midrule
      SWIRL (Ours) & $\max \{\lvert\theta_{i}\rvert\}$ \\
      \bottomrule
    \end{tabular}%
  }
\end{wraptable}Fig.~\ref{fig:varys_infer} presents the inference pipeline. Training alternates between the two agents: when optimizing the Navigator, the Interactor is fixed and executes the Navigator’s instructions to yield actions and rewards; when optimizing the Interactor, the Navigator remains fixed and supplies instructions for each step. This instantiation separates planning from execution and enforces a tight linkage between reasoning and action. With this decoupled design and the stability of interleaved updates, our system attains state-of-the-art zero-shot performance on both high-level and low-level mobile GUI benchmarks using only $3{,}500$ training examples, outperforming baselines trained under diverse SFT and RL regimes. We further apply the same interleaved recipe to a mathematics setting following prior multi-agent work~\citep{liao2025marft} and observe significant gains, including a $+14.8$ improvement on MATH500.


The contributions of this paper are as follows: (i) we introduce SWIRL, a multi-agent training framework that interleaves single-agent updates and transforms MARL to a sequence of single-agent RL problems; it achieves \(O(1)\) actor memory usage by loading only the currently updated agent, and accommodates heterogeneous model sizes, data schedules, and update budgets; (ii) we establish formal guarantees, including a per step safety bound, a monotonic improvement result across rounds, and convergence of returns; (iii) we instantiate SWIRL for mobile GUI control with a Navigator and an Interactor and, through extensive experiments, show stable training and state-of-the-art zero-shot performance, together with ablations that clarify when interleaving helps; and (iv) we demonstrate transferability by applying the same training recipe to a non-GUI domain (e.g., mathematics) and observe consistent gains on standard benchmarks, indicating potential for broader multi-agent applications.

\section{Related Work}

\paragraph{Reinforcement Learning for Mobile GUI Control}
Reinforcement learning (RL) has recently emerged as a promising paradigm for GUI tasks. Unlike supervised fine-tuning (SFT), which requires large-scale annotated data, RL can learn effective policies from comparatively fewer samples while exhibiting stronger generalization to new tasks~\citep{chu2025sft}. A number of recent studies have investigated RL for GUI grounding, with research directions ranging from reward function design to policy optimization~\citep{lu2025ui, luo2025gui, liu2025infigui, zhou2025gui, yuan2025enhancing, tang2025gui, lee2025reguide}. By contrast, applications of RL to more complex mobile control scenarios that involve executing coarse-grained natural language instructions remain relatively limited~\citep{luo2025gui}. Furthermore, existing work has predominantly adopted single-agent settings, leaving multi-agent approaches largely unexplored.

\paragraph{Multi-Agent Systems Based on Large Language Models}
A parallel line of research explores multi-agent systems powered by LLMs to address complex tasks~\citep{hu2025owl, xiao2024tradingagents, xiang2025towards, zhao2024electoral, wu2024autogen, zhao2024longagent, du2023improving}. These systems typically assign specialized roles such as debating, voting, or negotiation, thereby structuring interactions and facilitating coordination without training. To enhance robustness and mitigate distribution shift~\citep{han2024llm}, recent studies have proposed strategies including persuasion-aware training~\citep{stengel2024teaching} and iterative self-improvement via SFT~\citep{subramaniam2025multiagent, zhao2025sirius}. An important challenge is how to effectively train model cooperation when only limited training data is available~\citep{tran2025multi}.

\paragraph{Multi-Agent Reinforcement Learning} 
Works in Multi-Agent Reinforcement Learning (MARL) largely falls into value-based methods~\citep{sunehag2017value,rashid2020monotonic} and actor-critic approaches~\citep{chu2019multi,de2020independent}. A core challenge is non-stationarity, as one agent’s update changes others’ observations. Alternating optimization (e.g., A2PO~\citep{wang2023order}, HARL~\citep{zhong2024heterogeneous}) update agents sequentially at the step level but still face scalability issues~\citep{canese2021multi,tran2025multi}. MARFT~\citep{liao2025marft} combines MARL with LLMs for mathematical problem but suffers from gradient conflicts and parameter drift as the scale increases. It further argues that unifying MARL and LLMs is harder than addressing either alone, highlighting the need for scalable frameworks to integrate them efficiently.

\begin{figure*}[t!]
    \centering
    \includegraphics[width=1.\textwidth]{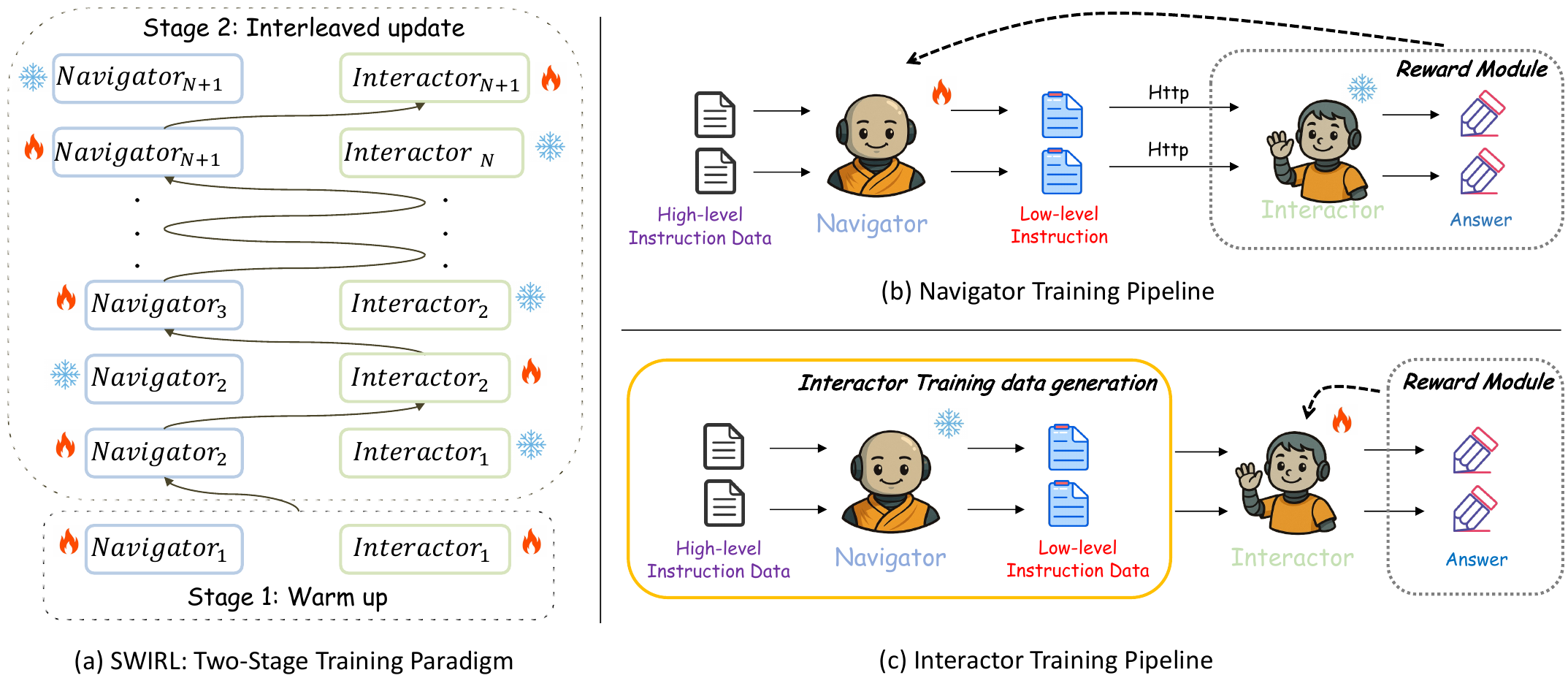}
    \caption{Our multi-agent training pipeline. (a) SWIRL decomposes multi-agent learning into two stages. Stage 1 performs warm-up, where each agent module (Navigator and Interactor) is initialized independently. Stage 2 proceeds with interleaved updates, where optimization alternates between agents: one module is updated while the other is frozen. (b) Given \textcolor{mypurple}{high-level instruction training data}, the Navigator generates \textcolor{red}{low-level instructions} and obtains rewards by making HTTP calls to a reward module that includes the Interactor (indicated by the dashed box). (c) First, the Navigator generates \textcolor{red}{training data (low-level instructions)} for the Interactor (indicated by the orange box), which are then used to train the Interactor.}
    \label{fig:method_overview}
\end{figure*}

\section{Method}
In Sec.~\ref{sec:alternating-updates}, we introduce a theoretical multi-agent interleaved updating methodology in Alg.~\ref{alg:blockwise} and give the theoretical guarantees. In Sec.~\ref{sec:task-formulation}, we formulate the multi-agent framework for GUI navigation. Following that, Sec.~\ref{sec:swirl} introduces SWIRL, a practical mobile GUI implementation of the multi-agent interleaved updating method, executed as a two-phase process involving warm-up and round-level alternating RL with online reweighting.

\subsection{Preliminary of Interleaved Updating}
\label{sec:alternating-updates}
The integration of MARL and LVLMs poses challenges, as each introduces unique complexities while being closely interlinked. Similar frictions have long been recognized in mathematical optimization. The Alternating Direction Method of Multipliers (ADMM)~\citep{boyd2011distributed} provides an effective strategy for tackling challenges formed by complex coupled problems: it decomposes an interconnected problem into manageable subproblems that are solved in turns. This divide-and-conquer approach is widely used in optimization, especially advantageous for challenging non-convex or non-differentiable objectives that can be decomposed into feasible subproblems~\citep{glowinski2014alternating,yang2022survey}. Consider the following constrained optimization problem $\min_{x, y\in\mathbb{R}^{n}} f(x) + g(y)
 \text{ s.t. } Ax + By = c$ and its augmented Lagrangian form: 
$
\mathcal{L}_\rho(x, y, \lambda) =
f(x) + g(y) + \lambda^\top (Ax + By - c)
+ \frac{\rho}{2} \|Ax + By - c\|^2$. To solve this, ADMM performs alternating optimization via the following iterative process: \(x^{k+1}:=\arg\min_x \mathcal{L}_\rho(x,y^k,\lambda^k); y^{k+1}:=\arg\min_y \mathcal{L}_\rho(x^{k+1},y,\lambda^k); \lambda^{k+1}:=\lambda^k+Ax^{k+1}+By^{k+1}-c\). It exemplifies problem decomposition and alternating optimization: rather than tackling a complex problem, alternating between simpler subproblems can achieve the overall objective.


\paragraph{From ADMM to Multi-Agent Interleaved Updating.}
In practical ADMM, each subproblem is solved by an inner loop to a prescribed accuracy before the outer iteration advances. The key is that these inner steps deliver enough improvement for the outer objective to make steady progress. Therefore, we propose a multi-agent round-level interleaved updating training scheme in Alg.~\ref{alg:blockwise}. The algorithm first independently pre-warms each agent using any single-agent method to obtain initial policies. It then performs interleaved updates: one agent is continuously optimized while the others are fixed, then sequentially switches to the next agent until all are updated, repeating this cycle to decompose MARL into a sequence of single-agent optimization tasks. This structure leads to the following findings: Proposition~\ref{prop:microstep-trpo} provides a safety bound for each micro-step, Theorem~\ref{thm:monotone} confirms that the return increases consistently across rounds, and Corollary~\ref{cor:ne} says the returns converge, and all policy limits attain this value. A summary of notation can be found in Appendix~\ref{app:Preliminary}, with complete proofs located in Appendix~\ref{app:Proofs Theoretical Results}.

\begin{proposition}[Lower bound at a micro-step]\label{prop:microstep-trpo}
In round $k$, if agent $\pi^{\,i}_{k,j}$ updates to $\pi^{\,i}_{k,j+1}$, the new joint policy is
$\Pi_{k,i,j+1} := (\tau^{-i}_{k},\ \pi^{\,i}_{k,j+1})$, and the performance satisfies
\begin{equation}\label{eq:P1}
J(\Pi_{k,i,j+1})\ \ge\ J(\Pi_{k,i,j})
\ +\ L^{\,i}_{\Pi_{k,i,j}}\!\big(\tau^{-i}_{k},\,\pi^{\,i}_{k,j+1}\big)-\ C_{k,i,j}\,D^{\max}_{\mathrm{KL}}\!\big(\pi^{\,i}_{k,j},\,\pi^{\,i}_{k,j+1}\big).
\end{equation}
\end{proposition}

\begin{algorithm}[t]
\caption{Multi-Agent Interleaved Updating with Monotonic Improvement Guarantee}
\label{alg:blockwise}
Initialise independent pre-warming $\pi_0^{i},\ 1\leq i\leq n$\;
\For{round $k = 0,1,2,\dots$}{
  \For{agent $i=1,\dots,n$}{
    Initialize $\pi^{\,i}_{k,0} \gets \pi_k^i$\;
    \For{micro-step $j=0,\dots,K_i-1$}{
      $\pi^{\,i}_{k,j+1} \gets \arg\max_{\pi^i}F_{k,i,j}(\pi^{\,i})=\big[L^{\,i}_{\Pi_{k,i,j}}(\tau^{-i}_{k},\pi^i)
      - C_{k,i,j}D_{\mathrm{KL}}^{\max}(\pi^{\,i}_{k,j},\pi^i)\big]$\;
    }
    Update $\pi^i_{k+1} \gets \pi^{\,i}_{k,K_i}$\;
  }
}
\end{algorithm}

\begin{theorem}[Monotonic improvement]\label{thm:monotone}
Every micro-step updates in Alg.~\ref{alg:blockwise} satisfies $F_{k,i,j}(\pi^{\,i}_{k,j+1})\ge 0$, and for all outer rounds $k$ we have $J(\pi_{k+1}) \ge J(\pi_k)$.
\end{theorem}

%

\begin{corollary}[Return convergence]\label{cor:ne}
The sequence $\{J(\pi_k)\}$ approaches a limit, referred to as $\bar J$, and the collection of limit points from $\{\pi_k\}$ is non-empty. For any subsequence $\{\pi_{k_j}\}_{j\ge0}$ that converges such that $\pi_{k_j}\rightarrow\bar\pi$, it holds that $J(\bar\pi) = \bar J$.
\end{corollary}

Alg.~\ref{alg:blockwise} provides a methodology: in each round, a single active agent performs several micro-steps while all other agents are frozen. Since the baseline stays constant throughout these micro-steps, each step simplifies to a typical single-agent policy update. As a result, the surrogate goal \(L - D_{\mathrm{KL}}^{\max}\) can be estimated using well-established single-agent techniques like TRPO~\citep{schulman2015trust}, PPO~\citep{schulman2017proximal}, and GRPO~\citep{shao2024deepseekmath}, which apply feasible trust region or clipped KL updates. Previous studies have verified both the theoretical validity and empirical effectiveness of this approximation~\citep{schulman2015trust,zhong2024heterogeneous}. We then apply this methodology to the GUI navigation task.

\subsection{Multi-Agent Framework for GUI Control}
\label{sec:task-formulation}
\paragraph{Task Formulation.}
We formulate GUI control task as a sequential decision-making problem. With a natural language instruction $I$, the agent reviews a series of historical screenshots and actions $H_{t}=\{X_{t-\delta_{s}}, \dots, X_{t-1},r^a_{t-\delta_{a}}, \dots, r^a_{t-1}\}$ along with the current screenshot $X_{t}$ to craft a structured text reply $r_t$ at each time step $t$. Here, $\delta_{s}$ and $\delta_{a}$ denote the counts of past screenshots and actions, respectively. This reply includes a reasoning ($r^r_t$) and a low-level instruction (LLI, $r^i_t$) that outlines the next planned step. The LLI is then succeeded by the actual action ($r^a_t$), where $r_t \sim \pi_{\theta_n, \theta_i}(r \mid I,H_{t},X_{t}),\ \{r^r_t,r^i_t,r^a_t\} \in r_t.$
The objective is to generate the next action $r^a_t$ that complies with the given instruction $I$. Appendix~\ref{apdx:examples} presents illustrative examples of GUI agents completing GUI control tasks. 
Nevertheless, navigating GUI-based instructions introduces specific challenges: it necessitates high-level navigation to deduce the next-step instruction and detailed perception to engage with UI components, each demanding distinct skills.

\paragraph{Architecture and Training Objective.}
To proficiently manage the complexities of GUI operations, we utilize a multi-agent system that distinctly separates the responsibilities between the \emph{Navigator} ($\pi_{\theta_n}$) and the \emph{Interactor} ($\pi_{\theta_i}$). The \emph{Navigator} is responsible for {high-level planning}, where it interprets the natural language instructions, merges past actions with the UI views, and establishes a coherent task context with reasoning. It then creates a detailed LLI, reflecting the intended next actions, based on the reasoning process. Subsequently, the \emph{Interactor} combines the LLI and the current UI view to generate concrete atomic actions, including actions like {click}, {scroll}, etc. This involves precise cursor positioning and visual interpretation to ensure accurate execution of the planned steps within the interface. The inference pipeline of the system is illustrated in Fig.~\ref{fig:varys_infer}.

Building on the system architecture above, we train the two agents with a round-level interleaved scheme (Alg.~\ref{alg:blockwise}). In each round, we select one role, either the Navigator or the Interactor, and run multiple inner updates on its parameters while freezing the other agent. We then swap roles and repeat. Optimizing the theoretical update for an individual agent demands the calculation of advantage \(A\), surrogate \(L\), and \(D_{\mathrm{KL}}^{\max}\), which is cost-prohibitive. Consequently, we implement practical relaxations, similar to~\citep{zhong2024heterogeneous}, by approximating the theoretical goal using a GRPO objective~\citep{shao2024deepseekmath}. Concretely, we calculate single-agent improvements using \emph{group-relative advantages}, denoted as \(A_k\), which are derived from multiple rollouts (standardized across the batch). We control intractable $D_{\mathrm{KL}}^{\max}$ with two tractable terms: clipped ratios around $\pi_{\text{old}}$ and a KL anchor to $\pi_{\text{ref}}$, ensuring local trust-region control and curbing drift for stable. This preserves the ascent direction of the theoretical target \(L - D_{\mathrm{KL}}^{\max}\), whilst ensuring stability and efficiency. Each agent’s action is an autoregressive sequence, responses for rollouts $K$ are \(\{r_{k,\ell}\}_{1\le k\le K}^{1\le \ell\le |r_k|}\). We compute token-wise importance ratios aligned with GRPO, in line with~\citep{luo2025gui}. Each token is assigned to either the navigator or interactor by the indicator \(\mathbb{I}^{(j)}_{k,\ell}=1\) if and only if \(r_{k,\ell}\sim \pi_{\theta_j}\) (and \(0\) otherwise), securing agent-specific credit assignment without breaching the ``freeze-the-complement" rule. In conclusion, our overarching multi-agent training objective is:

\begin{equation}\label{eq:SWIRLonj}
    \mathcal{J}(\theta_n, \theta_i) =
\mathbb{E}_{\substack{r \sim \pi \\ (I,H_{t},X_{t})\sim\mathcal{D}}} \left[ 
\sum_{k,\ell} \sum_{j \in \{n, i\}} \frac{\mathbb{I}^{(j)}_{k,\ell}}{K\sum_{\ell=1}^{|r_{k}|}\mathbb{I}^{(j)}_{k,\ell}}
 \cdot( 
\operatorname{clip}(v^{(j)}_{k,\ell}, A_k) 
- \lambda D_{\mathrm{KL}}[\pi_{\theta_j} \| \pi^{\text{ref}}_{\theta_j}] )
\right].
\end{equation}

The clipped surrogate is: $\operatorname{clip}(v, A) = \min\left(v A,\, \text{clip}(v, 1{-}\epsilon, 1{+}\epsilon) A\right),$ where the value of importance ratio is: $v^{(j)}_{k,\ell} = \frac{
\pi_{\theta_j}(r_{k,\ell} \mid  I,H_{t},X_{t}, r_{k,<\ell})
}{
\pi^{\text{old}}_{\theta_j}(r_{k,\ell} \mid I,H_{t},X_{t}, r_{k,<\ell})
}.$ The scalar reward is composed of: $R_k = \alpha R_{\text{form}} + \beta R_{\text{acc}},$ where $R_{\text{form}}$ denotes the reward for format correctness (e.g., proper usage of required tags), and $R_{\text{acc}}$ is the reward for action accuracy, defined as $R_{\text{acc}} = \lambda_1 R_{\text{act}} + \lambda_2 R_{\text{info}}$, where $R_{\text{act}}$ measures the correctness of the predicted action type, and $R_{\text{info}}$ quantifies the accuracy of the action parameters (e.g., the click location). Finally, the normalized advantage is computed as $A_k = \frac{R_k - \mu}{\sigma}$, where $\mu$ and $\sigma$ are the mean and standard deviation of rewards across sampled trajectories.

\subsection{SWIRL: Staged Workflow for Interleaved Reinforcement Learning}
\label{sec:swirl}
Building on the multi-agent architecture and learning objective described above, we introduce \textbf{SWIRL}, a {Staged Workflow for Interleaved Reinforcement Learning}. This approach is crafted to efficiently coordinate and enhance the performance of both the Navigator and Interactor concurrently, as demonstrated in Fig.~\ref{fig:method_overview} and Alg.~\ref{alg:swirl}. The benefits of SWIRL are listed in Appendix.~\ref{appx:Benefits of SWIRL}.

\textbf{Stage 1: Warm-up initialization.} The Navigator is first initialized through lightweight Chain-of-Thought~\citep{wei2022chain} SFT, while the interactor is bootstrapped via initial reinforcement learning. The primary goal of this stage is to let each agent clearly learn its designated role: the navigator focuses on producing reasoning steps and LLI, whereas the interactor outputs the corresponding action. This warm-up phase establishes a robust foundation and minimizes variability at the outset.

\textbf{Stage 2: Interleaved update.} After warm-up, SWIRL progresses into a round-level alternating training stage. Within each round, one agent undergoes continuously optimization via reinforcement learning, while the other is kept static. This approach simplifies the intertwined multi-agent learning challenge into a series of static single-agent tasks. Therefore, it allows us to directly leverage modern single-agent RL algorithms, such as GRPO (see Alg.~\ref{alg:grpo}), thereby decreasing implementation complexity while preserving training stability and cooperative efficiency.  In the Navigator training process, the Navigator generates a reply $r = \{ r^{r}, r^i \}$ at each step. The frozen Interactor then executes $r^i$ to produce the final action $r^a$. Rewards are computed as described in Sec.~\ref{sec:task-formulation}, with the accuracy reward of $r^a$ serving as the Navigator's $R_{\text{acc}}$. To further enhance training efficiency and scalability, we deploy the Interactor on a separate server and integrate it within the reward module. The Navigator communicates with the Interactor via HTTP requests, enabling efficient reward computation during RL optimization. For Interactor training, we first use the frozen Navigator to generate low-level instructions for each training sample. The Interactor is then optimized via RL using these instructions as input, with rewards also computed according to Sec.~\ref{sec:task-formulation}. The training pipelines for the Navigator and Interactor are illustrated in 
Fig.~\ref{fig:method_overview}b and Fig.~\ref{fig:method_overview}c, respectively.

\textbf{Online reweighting.} After computing the GRPO-type advantages for each batch, we exclude those low-quality samples~\citep{meng2025mmeureka,cui2025process} by regulation $\mathcal{R}$. These low-quality samples typically arise due to collaborator mistakes, noise, or simplicity~\citep{shi2025towards}. As the model's performance increases, it might happen that the number of high-quality samples meeting our filtering standard becomes fewer than the batch size, leading to partially filled batches. To address this, we replenish the batch by randomly resampling from the remaining high-confidence instructions. This process produces reliable batches without introducing additional rollout cost, implicitly up-weights the informative samples to enhance training stability and convergence.

\begin{algorithm}[ht]
\caption{SWIRL: Staged Workflow for Interleaved Reinforcement Learning}
\label{alg:swirl}

\KwIn{Dataset $\mathcal{D}$, hyperparameters $N,\mu_{n},\mu_{i}$\;}

\vspace{0.3em}
\textbf{Stage 1: Warm-up initialization.}

Initialize navigator $\pi_{\theta_n}^{(1)}$ via supervised fine-tuning on instruction-action pairs\;

Initialize interactor $\pi_{\theta_i}^{(1)}$ via RL using fixed planner outputs\;

\vspace{0.3em}
\textbf{Stage 2: Interleaved reinforcement learning.}

\For{$k = 1$ \KwTo $N$}{
    \textbf{Navigator update:} freeze $\pi_{\theta_i}^{(k)}$, update $\pi_{\theta_n}$ using RL in Alg.~\ref{alg:grpo}:
    \[
    \theta_n^{(k+1)} \leftarrow RL(\theta_n \mid \mathcal{J},\theta_n^{(k)}, \theta_i^{(k)},\mu_{n});
    \]
    
    \textbf{Interactor update:} freeze $\pi_{\theta_n}^{(k+1)}$, update $\pi_{\theta_i}$ using RL in Alg.~\ref{alg:grpo}:
    \[
    \theta_i^{(k+1)} \leftarrow RL(\theta_i \mid \mathcal{J},\theta_n^{(k+1)}, \theta_i^{(k)},\mu_{i});
    \]
}
\end{algorithm}

\section{Experiment}

The experimental setup is described in Sec.~\ref{sec:exp-setup}, and the complete experimental details are provided in Appendix~\ref{apdx:exp-detail}. Sec.~\ref{sec:exp-high} and Sec.~\ref{sec:exp-low} present evaluations of SWIRL's zero-shot performance on high-level and low-level tasks, respectively. Sec.~\ref{sec:exp-math} investigates the proposed multi-agent training framework in the mathematics domain. Comprehensive ablation studies are reported in Appendix~\ref{apdx:ablation}.

\subsection{Settings}\label{sec:exp-setup}

\paragraph{Implementation Details.}
We use Qwen2.5-VL-3B~\citep{Qwen2.5-VL} as the base model for both the Navigator and Interactor agents. For historical context, we include only the sequence of past actions, excluding previous screenshots (i.e., $\delta_a = t-1$, $\delta_s = 0$). The weighting coefficients in the reward function in Sec.~\ref{sec:task-formulation} are set to $\alpha = 0.1$, $\beta = 0.9$, $\lambda_1 = 0.2$, and $\lambda_2 = 0.8$. For online reweighting, we define the rule $\mathcal{R}$ as:
$\mathcal{R} =
\begin{cases}
\text{keep}, & 0.1 < \overline{R_x} < 1,\\
\text{discard}, & \text{otherwise.}
\end{cases},$
where $\overline{R_x}$ denotes the average reward of sample $x$ across rollouts.

\paragraph{Training.}
We collected $1,500$ and $2,000$ mobile-control samples for Stage 1 and Stage 2, respectively (see Appendix~\ref{apdx:exp-detail-data} for details). In Stage 1 (warm-up), the Navigator is trained for $1$ epoch with SFT and the Interactor for $5$ epochs with GRPO~\citep{shao2024deepseekmath}, both using a learning rate of $1\times 10^{-6}$, batch size $32$, and DeepSpeed ZeRO-1 for the Navigator; the Interactor uses $5$ rollouts per sample.
In Stage 2 (SWIRL alternating updates), the two agents are alternately trained for $2$ epochs each per round over 20 rounds, maintaining the same hyperparameters; the Navigator uses $8$ rollouts per sample and the Interactor $5$. Stage 1 runs on $8\times$ NVIDIA A800 GPUs; Stage 2 uses $16\times$ A800s, with half for Interactor deployment as a vLLM-based~\citep{kwon2023efficient} inference service and half for training. The training framework builds on Qwen2.5-VL\footnote{\url{https://github.com/QwenLM/Qwen2.5-VL/tree/main/qwen-vl-finetune}}~\citep{Qwen2.5-VL} for SFT and VeRL\footnote{\url{https://github.com/volcengine/verl}}~\citep{sheng2025hybridflow} for RL.

\paragraph{Evaluation.}
We evaluate SWIRL on high-level tasks (AndroidControl-High~\citep{li2024effects}, GUIOdyssey~\citep{lu2024gui}) and low-level tasks (AndroidControl-Low~\citep{li2024effects}, GUI-Act~\citep{chen2024guicourse}, OmniAct~\citep{kapoor2024omniact}). Following the previous work~\citep{wu2024atlas, luo2025gui}, we report Type, GR, and SR in a zero-shot prompt setting to assess out-of-domain generalization. Appendix~\ref{apdx:exp-detail-eval} provides detailed information.
All evaluations are conducted in a zero-shot prompt setting to assess the models' out-of-domain generalization ability. 

\begin{table}[!t]
  \centering
  \caption{Performance comparison on the AndroidControl-High and GUIOdyssey datasets. \textbf{Bold} and \underline{underline} indicate the best and second-best performers, respectively.}
  \label{tab:main-result}
  \scalebox{0.9}{
  \begin{tabular}{l|c|ccc|ccc|c}
    \toprule
    \multirow{2}{*}{Models}  & \multirow{2}{*}{Method}  & \multicolumn{3}{c|}{AndroidControl-High}  & \multicolumn{3}{c|}{GUIOdyssey}  & \multirow{2}{*}{Overall}  \\
     & & Type & GR & SR & Type & GR & SR & \\
    \midrule
    GPT-4o   & Closed-Source & 63.06 & 30.90 & 21.17 & 37.50 & 14.17 &  5.36 & 28.69 \\
    OS-Atlas-4B & SFT     & 49.01 & 49.51 & 22.77 & 49.63 & 34.63 & 20.25 & 37.63 \\
    OS-Atlas-7B & SFT      & 57.44 & 54.90 & 29.83 & 60.42 & 39.74 & 26.96 & 44.88 \\
    UI-R1-3B & RFT         & 43.97 & 63.99 & 26.30 & 20.93 & 56.35 &  8.65 & 36.70 \\
    UI-R1-E-3B & RFT       & 29.67 & 61.50 & 14.37 &  7.59 & 62.87 &  1.94 & 29.66 \\
    GUI-R1-3B & RFT        & 58.04 & 56.24 & 46.55 & 54.84 & 41.52 & 41.33 & 49.75 \\
    GUI-R1-7B & RFT        & \textbf{71.63} & 65.56 & \textbf{51.67} & 65.49 & 43.64 & 38.79 & 56.13 \\
    ReGUIDE-7B & RFT       & -- & -- & 50.00 & -- & -- & -- & -- \\
    GPT-5 / UI-R1-3B  & Multi-Agent & 55.08 & 73.47 & 37.27 & 62.97 & 62.42 & 35.93 & 54.52  \\ 
    GPT-5 / UI-R1-E-3B & Multi-Agent & 60.36 & \textbf{74.79} & 40.65 & 65.28 & \underline{63.41} & 38.22 & 57.12 \\ 
    \rowcolor{green!12} GPT-5 / Interactor & Multi-Agent & 64.68 & \underline{74.62} & 49.53 & \underline{68.77} & 62.70 & \underline{44.21} & \underline{60.75} \\
    \midrule
    \rowcolor{varysorange!25}
    Navigator / Interactor & SWIRL  & \underline{66.72} & 71.19 & \underline{51.24} & \textbf{74.87} & \textbf{66.39} & \textbf{51.65} & \textbf{63.68}\\
    \bottomrule
  \end{tabular}
  }
\end{table}
\begin{table}[!t]
  \centering
  \caption{Performance comparison on web and desktop low-level tasks. The best and second-best in each column are indicated by \textbf{bold} and \underline{underline}, respectively.}
  \label{tab:main-low-other}
  \scalebox{0.9}{
  \begin{tabular}{l|ccc|ccc|ccc|c}
    \toprule
    \multirow{2}{*}{Models}  & \multicolumn{3}{c|}{GUI-Act-Web} & \multicolumn{3}{c|}{OmniAct-Web} & \multicolumn{3}{c|}{OmniAct-Desktop}  & \multirow{2}{*}{Overall}  \\
      & Type & GR & SR & Type & GR & SR & Type & GR & SR \\
    \midrule
    GPT-4o       & 77.09 & 45.02 & 41.84 & 79.33 & 42.79 & 34.06 & 79.97 & 63.25 & 50.67 & 57.11 \\
    OS-Atlas-4B  & 79.22 & 58.57 & 42.62 & 46.74 & 49.24 & 22.99 & 63.30 & 42.55 & 26.94 & 48.02 \\
    OS-Atlas-7B  & 86.95 & 75.61 & 57.02 & 85.63 & 69.35 & 59.15 & 90.24 & 62.87 & 56.73 & 71.51 \\
    UI-R1-3B     & 75.89 & 79.43 & 67.31 & 75.42 & 61.35 & 61.33 & 73.41 & 64.12 & 63.98 & 69.14 \\
    GUI-R1-3B    & 89.86 & 87.42 & 76.31 & 88.58 & 75.10 & 75.08 & 91.86 & \underline{78.37} & \underline{78.31} & 82.32 \\
    GUI-R1-7B    & \underline{90.85} & \textbf{88.06} & \underline{80.31} & \underline{91.16} & \underline{77.29} & \textbf{77.35} & \underline{92.20} & \textbf{83.36} & \textbf{83.33} & \underline{84.88} \\
    \midrule
    \rowcolor{green!12}
    \textbf{Interactor} &
    \textbf{95.00} & \underline{87.85} & \textbf{84.85} &
    \textbf{94.52} & \textbf{81.67} & \underline{77.32} &
    \textbf{94.65} & 77.09 & 72.97 & \textbf{85.10} \\
    \bottomrule
  \end{tabular}
  }
\end{table}

\subsection{High-Level Task Zero-shot Performance}\label{sec:exp-high}

We benchmark our approach against models trained under different paradigms and observe substantial gains. As shown in Table~\ref{tab:main-result}, our multi-agent framework with two 3B models achieves state-of-the-art zero-shot performance, surpassing the SFT-trained OS-Atlas-7B\citep{wu2024atlas} by $18.8$ points and the RFT-trained GUI-R1-7B~\citep{luo2025gui} by $7.55$ points.
When GPT-5~\citep{openai2025introducinggpt5} is used as a high-level to low-level planner for UI-R1-3B and UI-R1-E-3B~\citep{lu2025ui}, the scores increase by $17.82$ and $24.86$ points, 
confirming the advantage of separating planning from execution.  
Replacing GPT-5 with our Navigator trained using SWIRL and interleaved updates yields a further improvement of $2.93$ points, demonstrating the effectiveness of our training strategy in enhancing coordination and boosting downstream performance.

\begin{wraptable}{r}{0.4\linewidth} 
  \centering
  \caption{Performance comparison on the AndroidControl-Low dataset. \textbf{Bold} and \underline{underline} indicate the best and second-best performers, respectively.}
  \label{tab:main-ac}
  \scalebox{0.80}{
  \begin{tabular}{lccc}
    \toprule
    \multirow{2}{*}{Model} & \multicolumn{3}{c}{AndroidControl-Low} \\
     & Type & GR & SR \\
    \midrule
    GPT-4o      & 74.33 & 38.67 & 28.39 \\
    OS-Atlas-4B & 64.58 & 71.19 & 40.62 \\
    OS-Atlas-7B & 73.00 & 73.37 & 50.94 \\
     UI-R1        & 72.49 & 88.48 & 57.37 \\
     UI-R1-E      & 73.91 & \underline{91.91} & 55.58 \\
     GUI-R1-3B    & 83.58 & 81.59 & 64.41 \\
     GUI-R1-7B    & \textbf{85.17} & 84.02 & 66.52 \\
     ReGUIDE-7B   & --    & --    & 67.40 \\
     SE-GUI-7B & -- & 79.60 & \underline{68.20} \\
    \midrule
    \rowcolor{green!12} Interactor              & \underline{84.62} & \textbf{92.20} & \textbf{78.81} \\
    \bottomrule
  \end{tabular}}
\end{wraptable}
\subsection{Low-Level Task Zero-shot Performance}\label{sec:exp-low}
The Interactor trained with SWIRL interleaved updates can also operate independently as a low-level task executor.
Although its low-level instruction inputs during Stage 2 training are generated by the Navigator, it still achieves strong results, recording the highest SR and GR scores ($78.81$ and $92.20$) and the second-highest Type score ($84.62$) on the AndroidControl-Low benchmark~\citep{li2024effects}, as shown in Table~\ref{tab:main-ac}.
Notably, despite all $3{,}500$ training samples across both stages originating exclusively from mobile devices, the model achieves an overall score of $85.10$ on low-level tasks in the web- and desktop-based GUI-Act~\citep{chen2024guicourse} and OmniAct~\citep{kapoor2024omniact} datasets, representing the best average performance (Table~\ref{tab:main-low-other}). This remarkable cross-domain generalization underscores the robustness of our model and provides strong empirical evidence for the effectiveness of our training methodology.

\subsection{Multi-Agent Training Framework in the Mathematics Domain}\label{sec:exp-math}

We adapt SWIRL to the mathematics domain to validate the generalizability and transferability of our multi-agent training framework. In line with the prior framework MARFT~\citep{liao2025marft}, we employ Qwen2.5-Coder-3B-Instruct~\citep{hui2024qwen2} as the base model to construct a dual-agent architecture, which is then trained on the MATH~\citep{hendrycks2021measuring} training set. Implementation details are provided in Appendix~\ref{apdx:exp-detail-math}. 
As shown in Table~\ref{tab:math-result}, SWIRL delivers consistent improvements across all benchmarks, achieving a $14.8$-point gain over MARFT on the in-domain MATH500 test set, and further surpassing it on the cross-domain CMATH~\citep{wei2023cmath} ($83.5$ vs.\ $83.0$) and out-of-domain GSM8K~\citep{cobbe2021training} ($81.4$ vs.\ $78.7$) datasets.
These results demonstrate that SWIRL not only excels in its original GUI mobile control domain but also transfers effectively to substantially different problem settings, highlighting its robustness and broad applicability.

\begin{table}[!t]
  \centering
  \caption{Math results for training-time multi-agent frameworks. SWIRL and MARFT share the same base model Qwen2.5-Coder-3B-Instruct, and are both trained on the MATH training set.}
  \scalebox{0.85}{
  \label{tab:math-result}
  \begin{tabular}{l|cc|ccc|c}
    \toprule
    Model & Method & Training Data & MATH500 & CMATH & GSM8K & Overall \\
    \midrule
     Qwen2.5-Coder-3B-Instruct & Multi-Agent & -- & 47.2 & 81.1 & 77.3 & 68.5 \\
     MARFT & Multi-Agent & MATH & 49.8 & 83.0 & 78.7 & 70.5 \\
     SWIRL (ours) & Multi-Agent & MATH & \textbf{64.6} & \textbf{83.5} & \textbf{81.4} & \textbf{76.5} \\
    \bottomrule
  \end{tabular}
  }
\end{table}

\section{Conclusion}
In this paper, we presented SWIRL, an interleaved reinforcement learning paradigm that reformulates multi-agent training into tractable single-agent updates. The central principle of SWIRL lies in its round-level alternating training strategy, where in each round one agent is continuously optimized through reinforcement learning while the others remain fixed. We provided theoretical guarantees for stable optimization and validated the effectiveness of SWIRL through extensive experiments in both mobile GUI control and multi-agent mathematical reasoning. Looking forward, we hope that SWIRL can inspire new approaches to multi-agent training in broader domains, such as finance and AI for science, where efficient coordination and reliable optimization remain critical challenges.

\bibliography{iclr2026_conference}
\bibliographystyle{iclr2026_conference}

\newpage
\appendix
\section{Preliminary and Proofs}

\subsection{Preliminary with Notation}
\label{app:Preliminary}
\paragraph{Environment and policies.}

We consider a cooperative Markov game
$\langle \mathcal N,\mathcal S,\mathcal A, r, P, d\rangle$
with $n$ agents: $\mathcal N=[n]=\{1,\ldots,n\}$. Let
$\mathcal A:=\prod_{i\in\mathcal N}\mathcal A^i$ be the joint action space.
For each $i\in\mathcal N$, the (stochastic Markov) policy
$\pi^i(\cdot\mid s)\in\Delta(\mathcal A^i)$ together form the joint policy
$\pi=(\pi^i)_{i\in\mathcal N}$, which induces the joint action distribution
$\pi(a\mid s)=\prod_{i\in\mathcal N}\pi^i(a^i\mid s)$ for
$a=(a^i)_{i\in\mathcal N}$. Here
$\Delta(\cdot)$ denotes the set of probability distributions over a set. The environment transitions according to
$P:\mathcal S\times\mathcal A\to\Delta(\mathcal S)$, written
$s_{t+1}\sim P(\cdot\mid s_t,a_t)$, with initial state $s_0\sim d$ and joint reward
$r:\mathcal S\times\mathcal A\to\mathbb R$. Let $\rho_{\pi}$ be the average state-visitation distribution induced by the joint policy $\pi$. For agent-wise decomposition we can write
$a=(a^{-i},a^i)$ and denote $\pi^{-i}=(\pi^j)_{j\neq i}$. 

\paragraph{Joint return $J(\pi)$.}
For discount $\gamma\in[0,1)$ and any fixed initial-state distribution,
\[
J(\pi):=\mathbb{E}_{\pi}\!\left[\sum_{t=0}^{\infty}\gamma^t\,r(s_t,a_t)\right].
\]
It represents the expected total reward from this point forward under the policy $\pi$: summing the reward from each subsequent step and applying a discount factor of $\gamma^t$ at step $t$. The notation $\infty$ indicates “assess each future step”; in tasks with a finite horizon, this sum naturally concludes upon task completion. The discount factor $\gamma$ dictates the temporal range: $\gamma=0$ focuses only on immediate rewards, suitable for offline tasks, where $J(\pi)=\mathbb{E}_\pi[r(s_0,a_0)]$, whereas a higher value of $\gamma$ emphasizes rewards further in the future.

\paragraph{Rounds, order, and micro-steps.}
Outer rounds are indexed by $k=0,1,2,\ldots$. Agent $i$ executes a block of $K_i$ \emph{micro-steps} indexed by $j=0,\ldots,K_i-1$, and we denote a micro-step by $(k,i,j)$. The joint policy at the start (resp. end) of round $k$ is $\pi_k$ (resp. $\pi_{k+1}$). During agent $i$’s block, $\pi^{\,i}_{k,j}$ is its temporary policy after $j$ micro-steps, initialized by $\pi^{\,i}_{k,0}=\pi_k^i$; after finishing the block, set $\pi_{k+1}^i:=\pi^{\,i}_{k,K_i}$. Other agents are held fixed according to the rolling baseline defined next.

\paragraph{Rolling baseline and complement policy.}
 The baseline joint policy at micro-step $(k,i,j)$ is
\[
\Pi_{k,i,j}
:=\big(\{\pi^{\,r}_{k+1}\}_{r<i},\ \pi^{\,i}_{k,j},\ \{\pi^{\,r}_{k}\}_{r>i}\big)
= \big(\tau^{-i}_{k},\,\pi^{\,i}_{k,j}\big),
\]
where the \emph{complement policy} (all agents except $i$) is
\[
\tau^{-i}_{k}:=\big(\{\pi^{\,r}_{k+1}\}_{r<i},\ \{\pi^{\,r}_{k}\}_{r>i}\big).
\]
During micro-step $j$ of agent $i$ in the $k$th round, agents positioned earlier in the sequence ($r<i$) have already adopted their policies for round $(k\!+\!1)$, the current agent follows its internal iterate $\pi^{\,i}_{k,j}$, and those positioned later ($r>i$) continue to use their round-$k$ policies. 

\paragraph{Value, $Q$, and advantages.}
Define
\[
V_\pi(s):=\mathbb{E}_\pi\!\Big[\sum_{t\ge0}\gamma^t r(s_t,a_t)\,\Big|\,s_0=s\Big],\qquad
Q_\pi(s,a):=\mathbb{E}_\pi\!\Big[\sum_{t\ge0}\gamma^t r(s_t,a_t)\,\Big|\,s_0=s,\,a_0=a\Big].
\]
The joint advantage is $A_\pi(s,a):=Q_\pi(s,a)-V_\pi(s)$.

For agent $i$, define the marginal state–action value and the single-agent advantage by
\[
Q^i_\pi(s,a^i):=\mathbb{E}_{a^{-i}\sim \pi^{-i}(\cdot\mid s)}\big[\,Q_\pi\big(s,(a^{-i},a^i)\big)\,\big],\qquad
A^{\,i}_\pi(s,a^i):=Q^i_\pi(s,a^i)-V_\pi(s).
\]
$A^{\,i}_\pi(s,a^i)$ is the local (per-state/per-action) improvement if agent $i$ takes $a^i$ at $s$ while others act according to $\pi^{-i}$.
A basic identity we use is the zero-mean property
$\mathbb{E}_{a^i\sim \pi^i(\cdot\mid s)}\!\big[A^{\,i}_\pi(s,a^i)\big]=0$ for every $s$.

\paragraph{Surrogate improvement.}
Given the baseline $\Pi_{k,i,j}$, the complement $\tau^{-i}_{k}$, and a candidate policy $\hat\pi^i$ for agent $i$, define
\[
L^{\,i}_{\Pi_{k,i,j}}\!\big(\tau^{-i}_{k},\hat\pi^i\big)
:=\mathbb{E}_{\substack{s\sim \rho_{\Pi_{k,i,j}}\\ a^i\sim \hat\pi^i(\cdot\mid s)}}\Big[\,A^{\,i}_{\Pi_{k,i,j}}(s,a^i)\,\Big].
\]

$L$ aggregates the local signal $A^{\,i}$ into a policy-level surrogate by averaging. It satisfies the baseline-zero property
$L^{\,i}_{\Pi_{k,i,j}}\!\big(\tau^{-i}_{k},\pi^{\,i}_{k,j}\big)=0$,
because $\mathbb{E}_{a^i\sim \pi^{\,i}_{k,j}(\cdot\mid s)}[A^{\,i}_{\Pi_{k,i,j}}(s,a^i)]=0$ for all $s$.

\paragraph{Micro-step objective.}
Let $\varepsilon_{k,i,j}:=\max_{s,a}|A_{\Pi_{k,i,j}}(s,a)|$ and
$C_{k,i,j}:=\tfrac{4\gamma\,\varepsilon_{k,i,j}}{(1-\gamma)^2}$.
Define the max conditional KL by
$D^{\max}_{\mathrm{KL}}(\mu,\nu):=\sup_s \mathrm{KL}\big(\mu(\cdot\mid s)\,\|\,\nu(\cdot\mid s)\big)$,
and the per-micro-step surrogate
\[
F_{k,i,j}(\hat\pi^i)\ :=\ L^{\,i}_{\Pi_{k,i,j}}(\tau^{-i}_{k},\hat\pi^i)\ -\ C_{k,i,j}\,D^{\max}_{\mathrm{KL}}\!\big(\pi^{\,i}_{k,j},\hat\pi^i\big).
\]
$F_{k,i,j}$ scores a candidate update as “local surrogate improvement $L$ minus a KL safety penalty”. Taking the current iterate as the candidate reveals that it possesses the baseline-zero property: $ F_{k,i,j}(\pi^{\,i}_{k,j}) =  0.$ This is because $L$ equals zero due to the zero-mean property of the single-agent advantage under the policy $\pi^{\,i}_{k,j}$, and $D^{\max}_{\mathrm{KL}}(\mu,\mu) = 0$.

\subsection{Theoretical Proofs}
\label{app:Proofs Theoretical Results}
\textbf{Proposition~\ref{prop:microstep-trpo}.}
In round $k$, if agent $\pi^{\,i}_{k,j}$ updates to $\pi^{\,i}_{k,j+1}$, the new joint policy is
$\Pi_{k,i,j+1} := (\tau^{-i}_{k},\ \pi^{\,i}_{k,j+1})$, and the performance satisfies
\begin{equation}\label{eq:P2}
J(\Pi_{k,i,j+1})\ \ge\ J(\Pi_{k,i,j})
\ +\ L^{\,i}_{\Pi_{k,i,j}}\!\big(\tau^{-i}_{k},\,\pi^{\,i}_{k,j+1}\big)-\ C_{k,i,j}\,D^{\max}_{\mathrm{KL}}\!\big(\pi^{\,i}_{k,j},\,\pi^{\,i}_{k,j+1}\big).
\end{equation}

\paragraph{Proof of Proposition~\ref{prop:microstep-trpo}.}
Set $\text{old}=\Pi_{k,i,j}$ and $\text{new}=\Pi_{k,i,j+1}=(\tau^{-i}_{k},\,\pi^{\,i}_{k,i,j+1})$.
Apply Lemma~6 in~\citep{zhong2024heterogeneous} with $\pi=\text{old}$ and $\bar\pi=\text{new}$. Because only agent $i$ changes,
the multi-agent surrogate in Lemma~6 reduces to $L^{\,i}_{\Pi_{k,i,j}}(\tau^{-i}_{k},\,\pi^{\,i}_{k,j+1})$, and the max-conditional KL reduces to
$D^{\max}_{\mathrm{KL}}(\pi^{\,i}_{k,j},\,\pi^{\,i}_{k,j+1})$. With
$C_{k,i,j}=\tfrac{4\gamma\,\varepsilon_{k,i,j}}{(1-\gamma)^2}$ and $\varepsilon_{k,i,j}=\max_{s,a}|A_{\Pi_{k,i,j}}(s,a)|$,
this yields exactly inequality \ref{eq:P2}. \qed

\textbf{Theorem~\ref{thm:monotone}.}
Every micro-step updates in Alg.~\ref{alg:blockwise} satisfies $F_{k,i,j}(\pi^{\,i}_{k,j+1})\ge 0$, then for all outer rounds $k$ we have $J(\pi_{k+1}) \ge J(\pi_k)$.

\paragraph{Proof of Theorem~\ref{thm:monotone}.}
Leveraging the baseline-zero characteristic, \( F_{k,i,j}(\pi^{\,i}_{k,j})=0 \), it follows that \(\arg\max_{\pi^i} F_{k,i,j}(\pi^i) \geq 0\). Therefore, by Proposition~\ref{prop:microstep-trpo} and the update rule,
\[
J(\Pi_{k,i,j+1})-J(\Pi_{k,i,j})\ \ge\ F_{k,i,j}\!\big(\pi^{\,i}_{k,j+1}\big)\ \ge\ 0
\]
for every micro-step $(k,i,j)$. Summing these inequalities in order over all micro-steps within round $k$ gives
\[
J(\pi_{k+1})-J(\pi_k)
=\sum_{i=1}^{n}\sum_{j=0}^{K_{i}-1}\big[J(\Pi_{k,i,j+1})-J(\Pi_{k,i,j})\big]\ \ge\ 0,
\]
hence $J(\pi_{k+1})\ge J(\pi_k)$ for all $k$. \qed

\textbf{Corollary~\ref{cor:ne}.}
The sequence $\{J(\pi_k)\}$ has a limit, denoted as $\bar J$, and the set comprised of limit points of $\{\pi_k\}$ is not empty. Additionally, for any convergent subsequence $\{\pi_{k_j}\}_{j\ge0}$ :$\pi_{k_j}\!\to\!\bar\pi$,
$J(\bar\pi)
=
\bar J$.

\paragraph{Proof of Corollary~\ref{cor:ne}.}
By Theorem~\ref{thm:monotone}, the performance sequence $\{J(\pi_k)\}_{k\ge0}$ is nondecreasing. 
With discount $\gamma\in[0,1)$ and bounded rewards $|r|\le R_{\max}$, every policy satisfies 
$|J(\pi)|\le R_{\max}/(1-\gamma)$, so $\{J(\pi_k)\}$ is bounded above and converges to some $\bar J\in\mathbb{R}$. Furthermore, as in~\citep{zhong2024heterogeneous}, the sequence of policies is bounded, hence it admits a convergent subsequence by the Bolzano-Weierstrass Theorem. Therefore, the set of limit points of $\{\pi_k\}$ is nonempty. Let $(\pi_{k_j})_{j\ge0}$ be any subsequence converging to a limit policy $\bar\pi$. 
By continuity of $J$ in $\pi$, we have
\[
J(\bar\pi)
\;=\;
J\!\left(\lim_{j\to\infty}\pi_{k_j}\right)
\;=\;
\lim_{j\to\infty}J(\pi_{k_j})
\;=\;
\bar J.
\]

 \qed

\section{Details in Method}
\subsection{Illustration of Algorithm~\ref{alg:grpo}}
In Alg.~\ref{alg:grpo}, 
$RL(\theta \mid \mathcal{J}, \cdot,  \cdot, \mu)$ 
denotes GRPO-type single agent RL based on the objective $\mathcal{J}$ defined in equation~\ref{eq:SWIRLonj} and hyperparameter $\mu$, where the first argument $\theta$ is the parameter to be updated.  

- When updating the \textit{navigator}, the interactor parameters are frozen. 
In this case, the input $\theta_{n}$ serves as the initial setting for the navigator, whereas $\theta_{\text{frozen}}$ pertains to the fixed interactor:
\[
RL(\theta \mid \mathcal{J}, \theta_{n}, \theta_{\text{frozen}}, \mu).
\]

- When updating the \textit{interactor}, the navigator parameters are frozen. 
In this case, the input $\theta_{i}$ serves as the initial setting for the interactor, whereas $\theta_{\text{frozen}}$ pertains to the fixed navigator:
\[
RL(\theta \mid \mathcal{J}, \theta_{\text{frozen}}, \theta_{i}, \mu),
\]
Below, we take the case of freezing the interactor and updating the navigator as an example.

\label{sec:Details in Method}
\begin{algorithm}[H]
\caption{GRPO-Type policy optimization $RL({\theta} \mid \mathcal{J}, \theta_{n}, \theta_{\text{frozen}}, \mu)$}
\label{alg:grpo}

\tcp{Here we take the case of freezing the interactor and updating the navigator as an example.}

\KwIn{Dataset $\mathcal{D}$, current policy $\theta_{n}$, frozen model $\theta_{\text{frozen}}$, hyperparameters $\mu=(M,B,K)$\;}

Initialize $\theta \leftarrow \theta_{n}$\;

\For{$\text{iteration} = 1 \dots M$}{
    Initialize reference policy $\theta_{\text{ref}} \leftarrow \theta$\;

    \For{$\text{step} = 1 \dots B$}{
        Sample a batch $\mathcal{D}_{b}$ from $\mathcal{D}$\;
        Update the old policy $\theta_{\text{old}} \leftarrow \theta$\;

        Sample $K$ outputs $\{r_k\}_{k=1}^K \sim \pi_{\theta_{\text{old}},\theta_{\text{frozen}}}(\cdot \mid I,H_{t},X_{t})$\;
        Compute rewards $\{R_k\}_{k=1}^K$ via reward model\;
        Compute GRPO-type advantages: $A_k \leftarrow \frac{R_k - \mu}{\sigma}$ over the group\;

        \textbf{Online reweighting:} discard samples with low quality and resample to refill the batch\;

        Update the policy model by maximizing the objective: 
        $\theta \leftarrow \arg\max_{\theta} \mathcal{J}(\theta, \theta_{\text{frozen}})$\;
    }
}

\KwRet{Updated parameters $\theta$\;}
\end{algorithm}

\subsection{Benefits of SWIRL}
\label{appx:Benefits of SWIRL}
Different from MARL methods such as HAPPO~\citep{zhong2024heterogeneous} (Fig.~\ref{fig:compare of SWIRL}a), 
A2PO~\citep{wang2023order}, 
and MARFT~\citep{liao2025marft} (Fig.~\ref{fig:compare of SWIRL}b), 
our approach (Fig.~\ref{fig:compare of SWIRL}c) offers four key advantages:

\begin{enumerate}
    \item \textbf{Seamless compatibility.} 
    SWIRL reformulates complex MARL tasks into an alternating sequence of simpler single-agent reinforcement learning problems. 
    This design naturally integrates with modern distributed single-agent RL frameworks 
    (e.g., veRL~\citep{sheng2025hybridflow}, OPEN-RLHF~\citep{hu2024openrlhf}), 
    eliminating intrusive modifications to communication protocols and enabling rapid adaptation to future efficient RL frameworks. 
    Consequently, our divide-and-conquer alternating solution fundamentally resolves the integration challenges 
    between MARL and contemporary LVLM-based training pipelines.  

    \item \textbf{Resource-friendly scalability.} 
    Consider a system with \(N\) agents, each with model size \(|\theta_{i}|\). 
    Table~\ref{tab:params} quantitatively summarizes the number of actor parameters loaded on the device during training.  
    In practice, A2PO, HAPPO, and MARFT keep all agents resident on the training device for local rollout. Under this setting, these methods typically load all actor parameters in the training device during learning, leading to a total parameter size of $\sum_{i=1}^{N}\lvert\theta_{i}\rvert$ 
    and memory consumption that scales linearly with the number of agents \((O(N))\).  
    In contrast, SWIRL only loads the currently updated agent model locally, 
    while executing the remaining agents as ``Model-as-a-Service'' modules remotely.  
    As a result, the memory usage of the training device remains constant regardless of \(N\) (i.e., \(O(1)\)), 
    greatly reducing hardware requirements and improving scalability for large-scale multi-agent systems.  

    \item \textbf{Adaptation to heterogeneity.} 
    Our method allows agents to adopt diverse model architectures and training configurations, 
    such as different numbers of training steps or heterogeneous datasets. Because SWIRL decomposes multi-agent training into per-agent optimization, it enables online, agent-conditioned reweighting that adapts to each agent’s distribution by filtering samples that are low-quality for that agent. As a result, heterogeneous agents receive tailored, high-confidence training signals; experiments on Appendix~\ref{sec:Effectiveness Analysis of Online Reweighting} empirically confirms reduced uninformative or misleading samples and improved overall training performance.

    \item \textbf{Stability.} 
    The alternating update methodology efficiently addresses the non-stationary challenges present in concurrent multi-agent training, thereby minimizing policy misalignment and distributional shift, as shown in Appendix~\ref{sec:Co-evolution of Navigator and Interactor} and~\ref{sec:Effectiveness Analysis of Online Reweighting}. Experimental results in Fig.~\ref{fig:rounds} consistently demonstrate stable performance enhancements with each alternating phase, thus facilitating effective coordination among multiple agents.

\end{enumerate}

\section{Experiment Details}\label{apdx:exp-detail}
\subsection{Training Data}\label{apdx:exp-detail-data}

\paragraph{Training Data Collection.}
We construct our two-stage training dataset based on the AITW\citep{rawles2023android} (using the expanded version from AITZ\citep{zhang2024android}) and AMEX\citep{chai2024amex} (adopting the Aguvis variant\citep{xu2024aguvis}), as both provide richer low-level annotations. 
In Stage 1, we leverage Qwen2.5-VL-3B to generate $8$ rollouts for each low-level instruction, then select $1,500$ samples whose average reward falls within the range $[0.3, 0.4]$. These high-quality samples are used in two ways: (1) the low-level instruction data is employed as RL training data for the Interactor, and (2) the reasoning traces associated with each sample are used to construct Chain-of-Thought supervised fine-tuning data with high-level instructions for the Navigator.
In Stage 2, we utilize the multi-agent system trained in Stage 1 to generate $8$ rollouts for each high-level instruction, filtering for $2,000$ samples with an average reward below $0.6$ and variance more than $0.175$. Notably, samples in Stage 2 include only high-level instructions and do not contain additional semantic annotations.
In total, we curate $3,500$ high-quality training samples across both stages, supporting robust chain-of-thought reasoning and effective RL optimization for both agents.

\paragraph{Action Space.}
Following the action space design style of UI-Tars~\citep{qin2025ui}, we define dataset-specific action spaces for different GUI benchmarks, as shown in Table~\ref{tab:appendix-action-spaces}.

\begin{figure*}[h!]
    \centering
    \includegraphics[width=1.\textwidth]{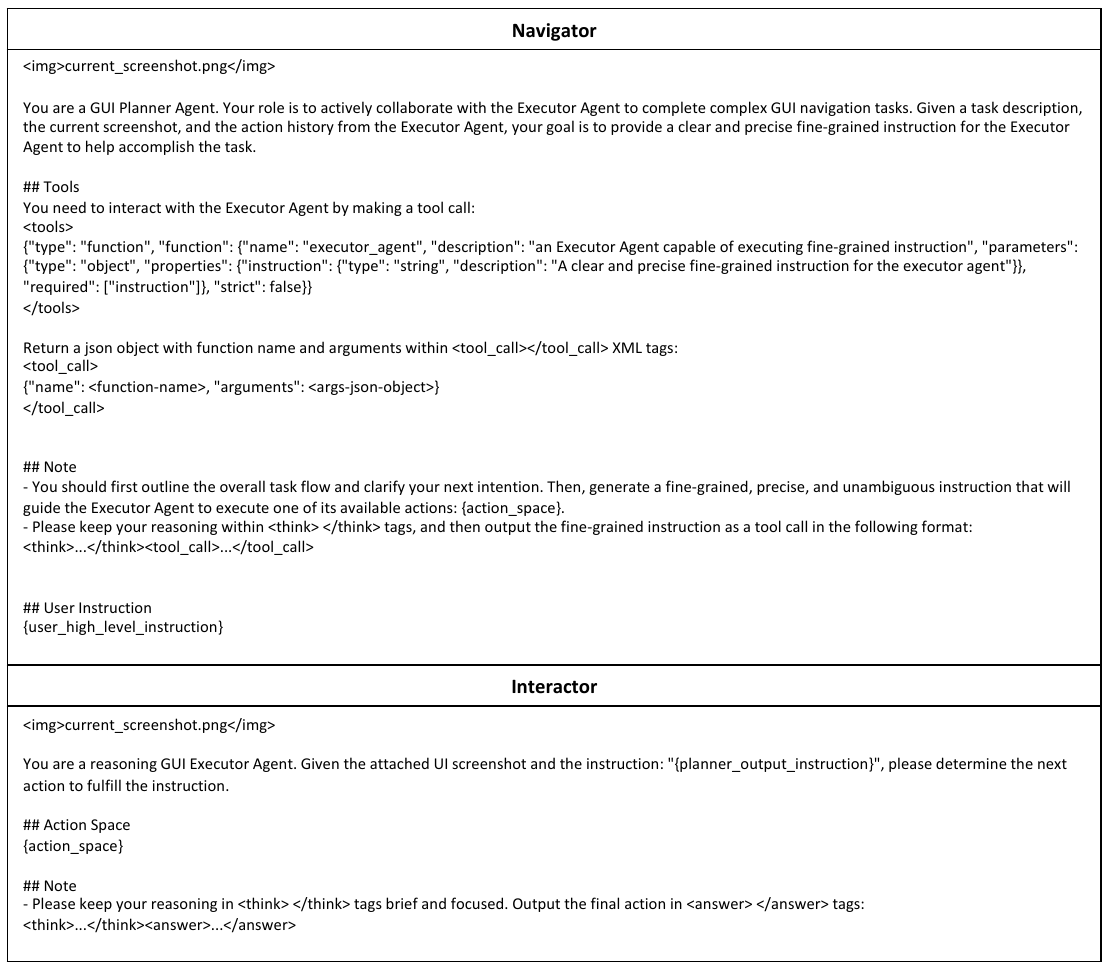}
    \caption{Prompt design for the dual-agent framework in the GUI domain.}
    \label{fig:prompt_gui}
\end{figure*}

\paragraph{Prompt.}
Fig.~\ref{fig:prompt_gui} illustrates the prompts used for the Navigator and Interactor.

\subsection{Evaluation Details}\label{apdx:exp-detail-eval}

\paragraph{Metrics.}
Following prior work~\citep{wu2024atlas, luo2025gui}, we evaluate our models using three standard metrics for GUI agents: Type, GR, and SR, which assess the accuracy of action type prediction, coordinate prediction, and step success rate, respectively. Type measures the exact match between the predicted action type (e.g., \texttt{click}, \texttt{scroll}) and the ground truth. GR evaluates the performance of GUI grounding in downstream tasks. SR (step-wise success rate) is computed by considering a step correct only if both the predicted action type and all associated arguments (e.g., coordinates for a click action) match the ground truth. 
For click-based actions (e.g., \texttt{click}, \texttt{long\_press}), the model must predict both the action type and the target coordinates (x, y). When the ground-truth bounding box is available, a prediction is considered correct if its coordinates fall within the box. If no bounding box is provided, or the prediction lies outside it, correctness is determined by whether the predicted coordinates are within $14\%$ of the screen width from the ground-truth position.
For type-based actions (e.g., \texttt{type}, \texttt{open\_app}), both the action type and content must match the ground truth. We compute the F1 score between the predicted and reference text, and an action is considered correct if $\text{F1} > 0.5$.
For scroll actions, the predicted direction argument (up, down, left, or right) must exactly match the ground truth.
For all other actions (e.g., \texttt{press\_enter}), correctness requires an exact match between the predicted action and the ground truth.

\paragraph{Baselines.} To ensure a fair comparison of cross-domain generalization, we select baseline models that have not been trained on the corresponding training sets of the evaluation domains. For high-level tasks, the comparison includes four categories of models: the proprietary GPT-4o\citep{gpt4o}; the state-of-the-art OS-Atlas-4B/7B\citep{wu2024atlas}, trained via supervised fine-tuning (SFT) on large-scale GUI grounding datasets; models trained with reinforcement fine-tuning (RFT) on GUI grounding datasets, including UI-R1-3B, UI-R1-E-3B\citep{lu2025ui}, GUI-R1-3B/7B\citep{luo2025gui}, and ReGUIDE-7B\citep{lee2025reguide}; and a multi-agent approach employing GPT-5\citep{openai2025introducinggpt5} as the planner to translate high-level instructions into low-level actions. For low-level tasks, the baselines consist of GPT-4o, OS-Atlas-4B/7B, and RFT-trained GUI grounding models, including UI-R1-3B, UI-R1-E-3B, GUI-R1-3B/7B, ReGUIDE-7B, and SE-GUI-7B\citep{yuan2025enhancing}.

\paragraph{Evaluation Datasets.}

\textbf{AndroidControl}\citep{li2024effects} is a mobile control dataset in which each GUI interaction trajectory is annotated with both coarse-grained high-level instructions and fine-grained low-level instructions, along with detailed XML metadata from which the bounding boxes of individual UI elements can be parsed. For click actions, we iterate over all bounding boxes in the XML, identify those containing the ground-truth coordinates, and select the smallest one by area as the candidate bounding box for click action evaluation.  
\textbf{GUIOdyssey}\citep{lu2024gui} consists of tasks involving cross-application operations, posing a significant challenge for models' planning abilities. In its latest release\footnote{\url{https://huggingface.co/datasets/hflqf88888/GUIOdyssey}}, each click action is supplemented with the bounding box of the corresponding UI element obtained via SAM2\citep{ravi2024sam} segmentation, which we also use for click action evaluation. Notably, the latest version contains $8{,}834$ samples, compared to $7{,}735$ in the earlier release\footnote{\url{https://huggingface.co/datasets/OpenGVLab/GUIOdyssey}}, with changes in the test set size. To ensure consistency with other baselines, we use the Test-Random split from the earlier version.  
\textbf{GUI-Act-Web}\citep{chen2024guicourse} contains web-based interaction data. To better assess GUI manipulation capabilities, we remove several QA-style samples from the test set and retain the remaining samples for evaluation. Bounding boxes are not used for click action evaluation in this dataset.  
\textbf{OmniAct}\citep{kapoor2024omniact} includes both web and desktop interaction data. Due to action space compatibility constraints, we filter out samples whose original action space involves hotkeys and keep the remaining samples for evaluation. Similarly, bounding boxes are not used for click action evaluation in this dataset.

\begin{table}[!t]
  \centering
  \caption{Action spaces used for each dataset.}
  \label{tab:appendix-action-spaces}
  \scalebox{0.98}{
  \begin{tabular}{l p{0.68\textwidth}}
    \toprule
    \textbf{Dataset} & \textbf{Action Space} \\
    \midrule
    AITW, AMEX &
      \makecell[l]{\texttt{click(point=`(x1, y1)')}\\
      \texttt{type(content=`xxx')}\\
      \texttt{scroll(direction=`down|up|right|left')}\\
      \texttt{press\_home()}\\
      \texttt{press\_back()}\\
      \texttt{press\_enter()}\\
      \texttt{finished()}} \\
    \midrule
    AndroidControl &
      \makecell[l]{\texttt{click(point=`(x1, y1)')}\\
      \texttt{long\_press(point=`(x1, y1)')}\\
      \texttt{type(content=`xxx')}\\
      \texttt{scroll(direction=`down|up|right|left')}\\
      \texttt{open\_app(app\_name=`xxx')}\\
      \texttt{press\_home()}\\
      \texttt{press\_back()}\\
      \texttt{wait()}\\
      \texttt{finished()}} \\
    \midrule
    GUIOdyssey  &
      \makecell[l]{\texttt{click(point=`(x1, y1)')}\\
      \texttt{long\_press(point=`(x1, y1)')}\\
      \texttt{type(content=`xxx')}\\
      \texttt{scroll(direction=`down|up|right|left')}\\
      \texttt{press\_home()}\\
      \texttt{press\_back()}\\
      \texttt{press\_appselect()}\\
      \texttt{error(content=`xxx')}\\
      \texttt{finished()}} \\
    \midrule
    GUI-Act-Web  &
      \makecell[l]{\texttt{click(point=`(x1, y1)')}\\
      \texttt{scroll(direction='down|up')}} \\
    \midrule
    OmniAct-Web) &
      \makecell[l]{\texttt{click(point=`(x1, y1)')}\\
      \texttt{rightclick(point=`(x1, y1)')}\\
      \texttt{scroll(direction=`down|up')}} \\
    \midrule
    OmniAct-Desktop &
      \makecell[l]{\texttt{click(point=`(x1, y1)')}\\
      \texttt{rightclick(point=`(x1, y1)')}\\
      \texttt{doubleclick(point=`(x1, y1)')}\\
      \texttt{moveto(point=`(x1, y1)')}\\
      \texttt{scroll(direction=`down|up')}} \\
    
    \bottomrule
  \end{tabular}
  }
\end{table}
\subsection{Mathematics Domains}\label{apdx:exp-detail-math}

\begin{figure*}[h!]
    \centering
    \includegraphics[width=1.\textwidth]{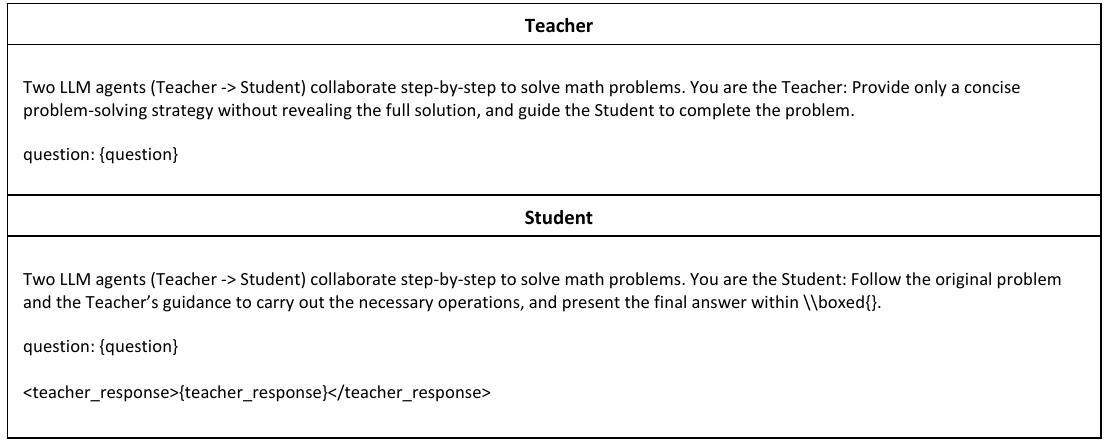}
    \caption{Prompt design for the dual-agent framework in the mathematics domain.}
    \label{fig:prompt_math}
\end{figure*}

\paragraph{Implementation Details.}
We design a dual-agent framework in which one agent acts as the \textit{Teacher}, providing a concise outline of the problem-solving approach, and the other acts as the \textit{Student}, generating the final solution by incorporating the Teacher's guidance. The detailed prompt formulations for both agents are provided in Fig.~\ref{fig:prompt_math}. Both agents are initialized from the Qwen2.5-Coder-3B-Instruct\citep{hui2024qwen2} model.
For the reward function, we assign a reward of $1$ if the generated answer is correct and $0$ otherwise, i.e.,
$R_x =
\begin{cases}
1, & \text{if answer is correct},\\
0, & \text{otherwise}.
\end{cases}$.
For online reweighting, we define the rule $\mathcal{R}$ as follows: we retain samples whose average reward across multiple rollouts lies strictly between $0.2$ and $0.8$, i.e.,
$\mathcal{R} =
\begin{cases}
\text{keep}, & 0.2 < \overline{R_x} < 0.8,\\
\text{discard}, & \text{otherwise.}
\end{cases},$
where $\overline{R_x}$ denotes the mean reward of sample $x$ over all rollouts.

\paragraph{Training.}
We train the dual-agent framework on the MATH\citep{hendrycks2021measuring} training set, which contains $7,500$ samples, by directly initiating Stage 2 of the SWIRL alternating-update procedure and bypassing the warm-up stage. In each round, the Teacher is updated first, followed by the Student, with each agent trained for $1$ epoch per round over a total of $10$ rounds. Both agents use a batch size of $128$ and a learning rate of $1\times 10^{-6}$ . The Teacher and Student perform $4$ and $8$ rollouts per sample, respectively. Training is conducted on $16\times$ NVIDIA A800 GPUs, with half allocated to deploying the Student as a vLLM-based~\citep{kwon2023efficient} inference service and the remaining half for model training.

\section{Ablation Study}\label{apdx:ablation}

\subsection{Effect of Interleaved Update}
To isolate the effect of SWIRL's second training stage (interleaved updates), we compare two models trained on the same total dataset of $3,500$ samples. Both start from the Stage 1 warm-up with $1,500$ samples. The first continues training on the remaining $2,000$ samples using the Stage 1 strategy until convergence, while the second proceeds with our Stage 2 alternating update scheme. The only difference between them is the training strategy applied in the second stage. As shown in Table \ref{tab:ablation-swirl-high}, although extended warm-up training with more data yields performance gains, its upper bound remains lower than that achieved by our interleaved update approach ($61.69$ vs. $63.68$), highlighting the latter's effectiveness.

\begin{table}[!h]
  \centering
  \caption{Effect of interactive updates. Numbers in \emph{italics} indicate performance gains relative to Stage 1 with $1,500$ samples.}
  \label{tab:ablation-swirl-high}
  \scalebox{0.7}{
  \begin{tabular}{c|ccc|ccc|c}
    \toprule
    \multirow{2}{*}{Training Strategy} & \multicolumn{3}{c|}{AndroidControl-High} & \multicolumn{3}{c|}{GUIOdyssey} & \multirow{2}{*}{Overall} \\
     & Type & GR & SR & Type & GR & SR & \\
    \midrule
    Stage 1 $+$ Stage 1
      & 63.77 \emph{(+0.90)} & 70.25 \emph{(+0.35)}  & 48.04 \emph{(+1.21)}  & 72.25 \emph{(+1.52)}  & 66.19 \emph{(+2.87)}   & 49.64 \emph{(+3.22)}  & 61.69 \emph{(+1.68)} \\
    Stage 1 $+$ Stage 2
      & \textbf{66.72 \emph{(+3.85)}}   & \textbf{71.19 \emph{(+1.29)}}    & \textbf{51.24 \emph{(+4.41)}}    & \textbf{74.87 \emph{(+4.14)}}   & \textbf{66.39\emph{(+3.07)} }    & \textbf{51.65\emph{(+5.23)} }   & \textbf{63.68 \emph{(+3.67)}}  \\
    \bottomrule
  \end{tabular}
  }
\end{table}

\subsection{Co-evolution of Navigator and Interactor}
\label{sec:Co-evolution of Navigator and Interactor}
We investigate the impact of the proposed interleaved update on individual agents within the multi-agent framework. For the Interactor, Table~\ref{tab:ablation-single-low} compares its performance on low-level tasks with and without Stage 2 interleaved updates. The results show a substantial improvement of $4.89$ points with Stage 2, particularly in SR (a metric that directly measures the correctness of individual actions and thus serves as a more critical indicator of the Interactor’s precision), which increases by nearly $7$ points.
For the Navigator, direct evaluation is less straightforward because its outputs are low-level instructions rather than executable actions. To address this, we adopt an indirect evaluation approach: we feed the Navigator's outputs into a fully trained Interactor and assess the resulting task performance. As shown in Table~\ref{tab:ablation-single-high}, pairing the Interactor with a Navigator trained using Stage 2 interleaved updates delivers consistently higher performance on high-level tasks, raising the overall score from $56.32$ to $63.68$ and yielding notable improvements in both GR and SR across benchmarks. This demonstrates that the updated Navigator produces more detailed and accurate low-level instructions from high-level goals.
Overall, these findings provide strong evidence that interleaved updates effectively enhance each agent's ability to fulfill its specific role, enabling them to co-evolve and achieve better collaborative performance within the multi-agent system.

\begin{table}[!h]
  \centering
  \caption{Effect of interleaved updates on Interactor performance in low-level tasks.
  }
  \label{tab:ablation-single-low}
  \scalebox{0.72}{
  \begin{tabular}{c|ccc|ccc|ccc|ccc|c}
  \toprule
  \multirow{2}{*}{Training Strategy}  & \multicolumn{3}{c|}{GUI-Act-Web} & \multicolumn{3}{c|}{OmniAct-Web} & \multicolumn{3}{c|}{OmniAct-Desktop}  & \multicolumn{3}{c|}{AndroidControl-Low}   & \multirow{2}{*}{Overall}  \\
      & Type & GR & SR & Type & GR & SR & Type & GR & SR & Type & GR & SR \\
    \midrule
    Stage 1  & 94.31 & \textbf{88.22} & 77.00 & 89.22 & 81.62 & 72.40 & 91.38 & 74.45 & 68.03 & \textbf{85.48} & 71.91 & 68.87 & 80.24 \\
    
    Stage 1 $+$ Stage 2 & \textbf{95.00} & 87.85 & \textbf{84.85} & \textbf{94.52} & \textbf{81.67} & \textbf{77.32} & \textbf{94.65} & \textbf{77.09} & \textbf{72.97} & 84.62 & \textbf{92.20} & \textbf{78.81} & \textbf{85.13} \\
    \bottomrule
  \end{tabular}
  }
\end{table}

\begin{table}[!h]
  \centering
  \caption{Performance of multi-Agent systems on high-level tasks with Navigators trained with and without interleaved updates.}
  \label{tab:ablation-single-high}
  \scalebox{0.9}{
  \begin{tabular}{c|ccc|ccc|c}
    \toprule
    \multirow{2}{*}{Stage 2 Training} & \multicolumn{3}{c|}{AndroidControl-High} & \multicolumn{3}{c|}{GUIOdyssey} & \multirow{2}{*}{Overall} \\
     & Type & GR & SR & Type & GR & SR & \\
    \midrule
    \ding{55} & 63.16 & 54.65 & 39.91 & 73.19 & 60.49 & 46.51 & 56.32 \\
    \ding{51} & \textbf{66.72 }   & \textbf{71.19}    & \textbf{51.24 }    & \textbf{74.87}   & \textbf{66.39 }    & \textbf{51.65 }   & \textbf{63.68}  \\
    \bottomrule
  \end{tabular}
  }
\end{table}

\subsection{Stability and Potential}
\begin{figure*}[t!]
    \centering
    \includegraphics[width=0.75\textwidth]{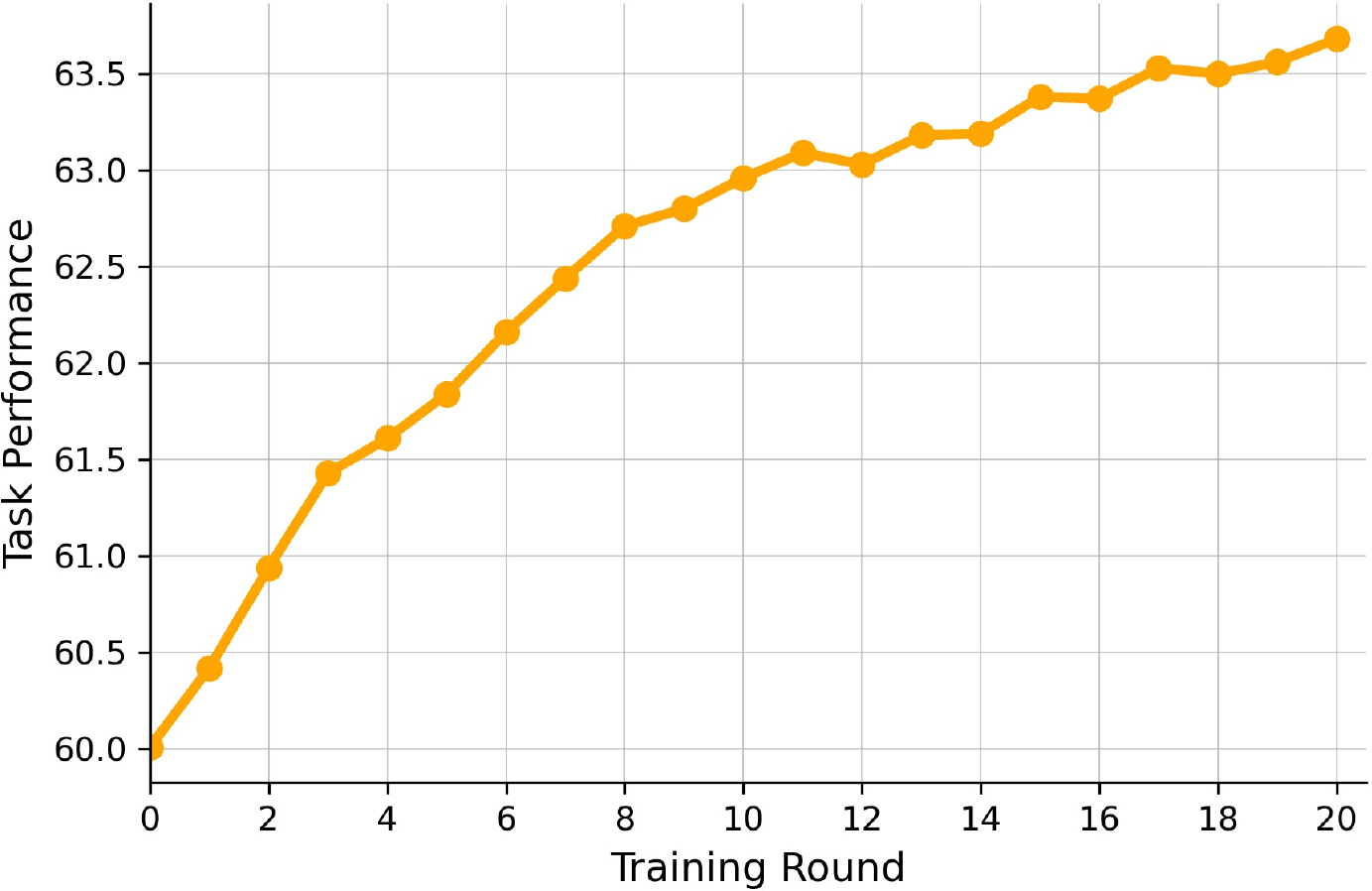}
    \caption{SWIRL training dynamics showing steady performance gains and stability across rounds. Task performance is measured as the average score across all high-level benchmarks, including AndroidControl-High and GUIOdyssey.}
    \label{fig:rounds}
\end{figure*}
As illustrated in Fig.~\ref{fig:rounds}, even with only $2,000$ samples used during the SWIRL's stage 2, the proposed training paradigm consistently enhances the model's generalization capability. 
The model's performance improves steadily with each training round, demonstrating the stability of the alternating optimization process. Furthermore, its out-of-domain generalization continues to increase in the later rounds, suggesting that the performance upper bound has not yet been reached. These results highlight the effectiveness and scalability of SWIRL for robust multi-agent training in GUI navigation tasks.

\subsection{Synergy and Individual Competence in Multi-Agent Training}
\begin{figure*}[h!]
    \centering
    \includegraphics[width=1.\textwidth]{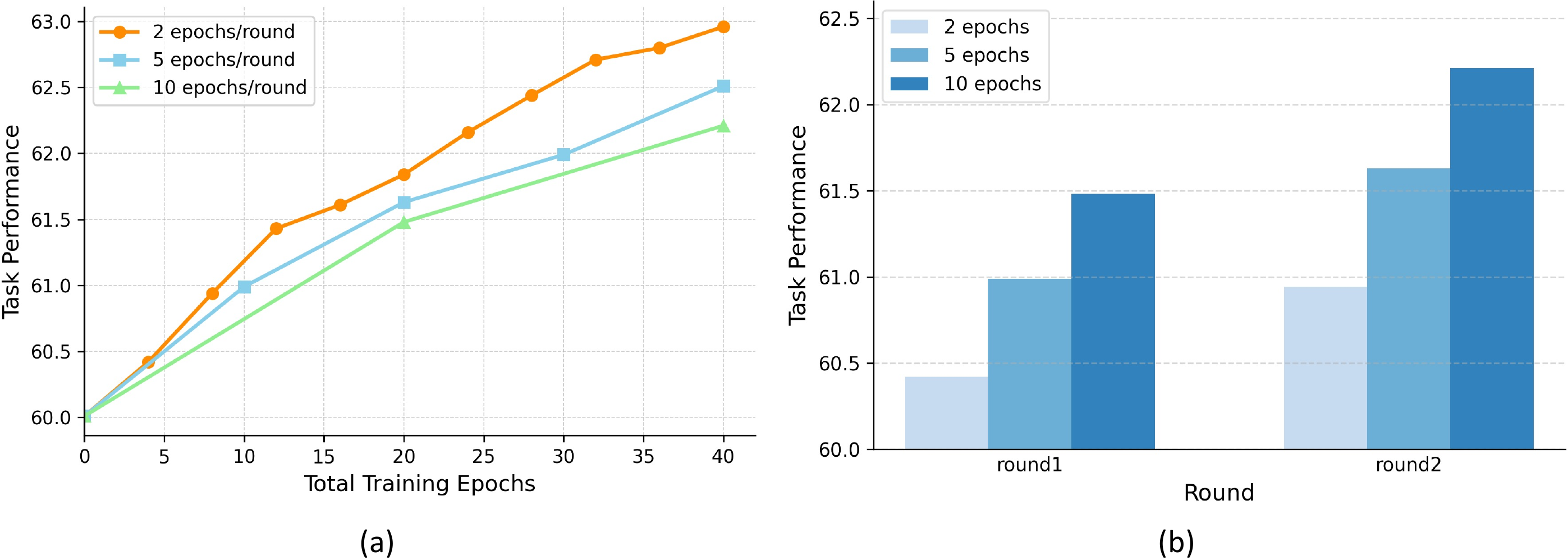}
    \caption{Effect of the training schedule on SWIRL performance with varying numbers of rounds and epochs per round. (a) Training curves under a fixed total number of training epochs, comparing different numbers of epochs per round. (b) Performance comparison after each round for different epochs-per-round settings.}
    \label{fig:vis3}
\end{figure*}

In multi-agent training, overall system performance depends not only on enhancing the capabilities of individual agents but also on achieving effective coordination among them. To disentangle and quantify the relative contributions of these two factors, we vary the number of alternating training rounds (i.e., the frequency and intensity of inter-agent coordination) and the number of epochs per round (i.e., the depth of single-agent training), while keeping the total number of training epochs constant. As shown in Fig.\ref{fig:vis3}(a) and Table~\ref{tab:rounds-epochs}, increasing the number of alternating rounds, thereby providing more opportunities for inter-agent interaction and iterative policy refinement, consistently leads to greater performance gains than allocating the same budget to deeper single-agent training within fewer rounds. At the same time, Fig.~\ref{fig:vis3}(b) indicates that increasing the training depth per round still brings incremental benefits, demonstrating that single-agent optimization remains valuable. Taken together, these results suggest that, under a fixed training budget, inter-agent coordination is the primary driver of performance improvement, while individual agent refinement plays a complementary yet meaningful role.

\begin{table}[!h]
  \centering
  \caption{Effect of the rounds-epochs schedule under a fixed training budget. Here, `Epochs/round' denotes the number of epochs trained within each round. The total training epochs are held constant across settings.}
  \scalebox{1}{
  \label{tab:rounds-epochs}
  \begin{tabular}{cc|ccc|ccc|c}
    \toprule
    \multirow{2}{*}{Rounds} & \multirow{2}{*}{Epochs/round}  & \multicolumn{3}{c|}{AndroidControl} & \multicolumn{3}{c|}{GUIOdyssey} & \multirow{2}{*}{Overall} \\
     & & Type & GR & SR & Type & GR & SR & \\
    \midrule
    2 & 10 & 65.74 & 70.42 & 49.88 & 73.13 & 64.85 & 49.27 & 62.21 \\
    5 & 4  & 65.26 & \textbf{70.94} & 49.81 & 73.85 & 65.21 & 49.98 & 62.51 \\
    10 & 2  & \textbf{66.00} & 70.91 & \textbf{50.56} & \textbf{74.03} & \textbf{65.71} & \textbf{50.58} & \textbf{62.96} \\
    
    \bottomrule
  \end{tabular}
  }
\end{table}

\subsection{Effectiveness Analysis of Online Reweighting}
\label{sec:Effectiveness Analysis of Online Reweighting}
\begin{figure*}[h!]
    \centering
    \includegraphics[width=1.\textwidth]{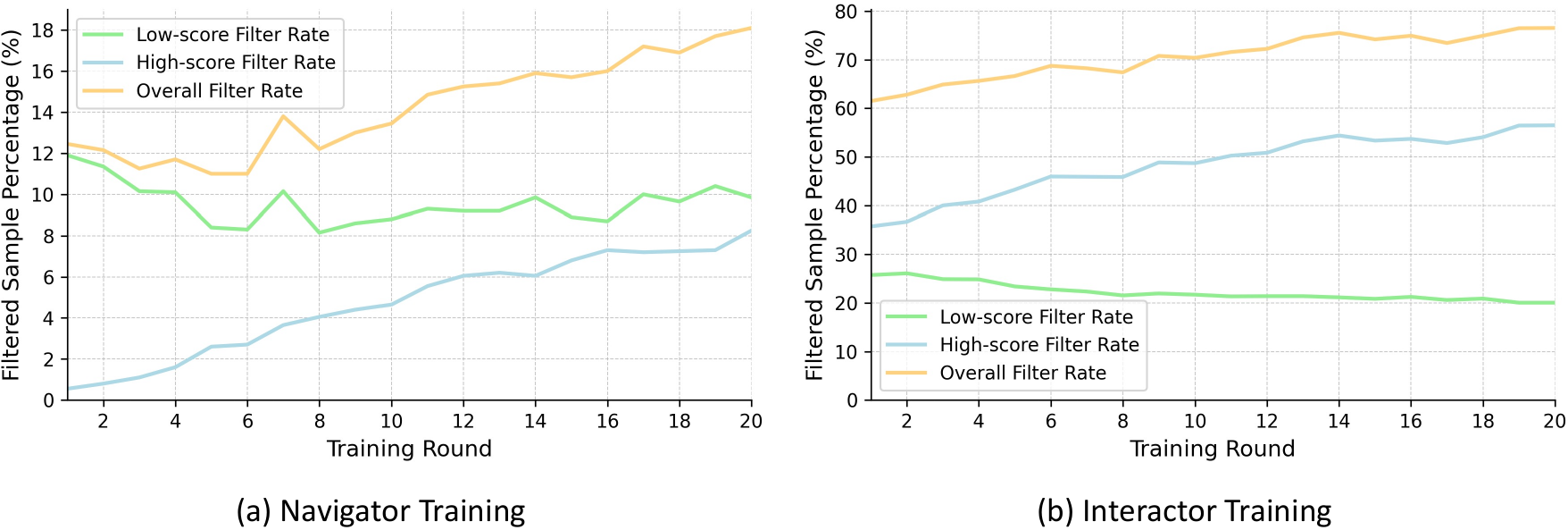}
    \caption{Filtered sample rates with online reweighting in SWIRL.}
    \label{fig:vis2}
\end{figure*}

To assess the impact of the online reweighting mechanism, we conduct comparative experiments with and without weighted resampling. As shown in Fig.~\ref{fig:ablation}(a), models trained without online reweighting perform significantly worse and even exhibit a degradation trend, underscoring the necessity of this mechanism. We hypothesize that online reweighting dynamically prioritizes informative samples, ensuring that they exert greater influence during training, which is essential for the stable improvement of multi-agent systems.
Further analysis, as illustrated in Fig.~\ref{fig:vis2}, reveals two important findings. First, the Interactor filters out approximately four times as many samples as the Navigator ($\sim$$75$ vs. $\sim$$18$), indicating a substantial difference in how the same dataset contributes to the learning process of each agent. The online reweighting strategy thus adapts to the evolving capabilities of each agent, assigning dynamic weights to those training samples most beneficial at each stage. Second, it is important to recognize that not all samples filtered for being completely incorrect are attributable solely to the deficiencies of a single agent. Some errors may arise from noisy data or from mistakes originating with another agent (e.g., if the Planner generates an erroneous instruction, the Executor may repeatedly fail to execute the correct action). In these cases, the online reweighting mechanism helps to exclude uninformative or misleading samples, thereby further improving the robustness and effectiveness of the training process.

\subsection{Sequential Updates versus Parallel Updates}
\begin{figure*}[h!]
    \centering
    \includegraphics[width=0.9\textwidth]{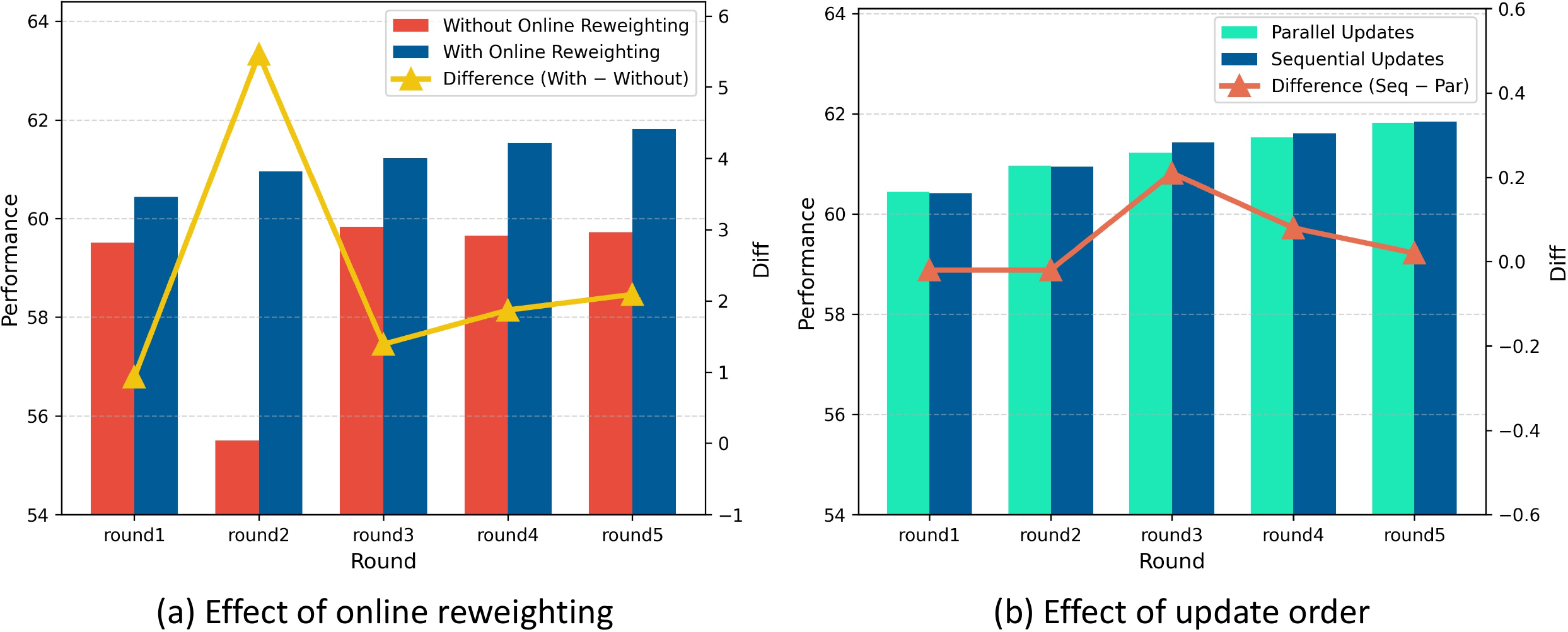}
    \caption{Ablation results on SWIRL zero-shot performance for AndroidControl-High and GUIOdyssey. (a) Online reweighting vs. no reweighting. (b) Sequential vs. parallel updates. Performance differences are indicated by lines.}
    \label{fig:ablation}
\end{figure*}

Our default SWIRL implementation employs a strictly sequential update scheme, in which the Navigator is updated first in each training round, followed by the Interactor. We also explore a parallel update strategy, where both agents are updated simultaneously within each round, rather than following a fixed order, i.e.,
$\left\{
\substack{
\theta_n^{(k+1)} \leftarrow RL({\theta_n}| \mathcal{J},\theta_n^{(k)}, \theta_i^{(k)},\mu_{n})\\
\theta_i^{(k+1)} \leftarrow RL({\theta_i}| \mathcal{J},\theta_n^{(k)}, \theta_i^{(k)},\mu_{i})
}
\right.$.
This approach aligns with the update mechanism of Jacobian ADMM~\citep{yang2022survey}, which relaxes the sequential constraints in standard ADMM and allows for simultaneous updates, i.e.,
$\left\{
\substack{
x^{k+1} := \arg\min_x \mathcal{L}_\rho(x, y^k, \lambda^k)\\
y^{k+1} := \arg\min_y \mathcal{L}_\rho(x^{k}, y, \lambda^k)
}
\right.$, thereby increasing training efficiency.
As shown in Fig.~\ref{fig:ablation}(b), parallel updates can achieve performance comparable to or surpassing that of sequential updates, even without strict alternation. This finding suggests that relaxing the sequential constraint in alternating multi-agent training can improve efficiency without sacrificing model quality.

\clearpage
\section{Examples}\label{apdx:examples}

\begin{figure*}[h!]
    \centering
    \includegraphics[width=0.65\textwidth]{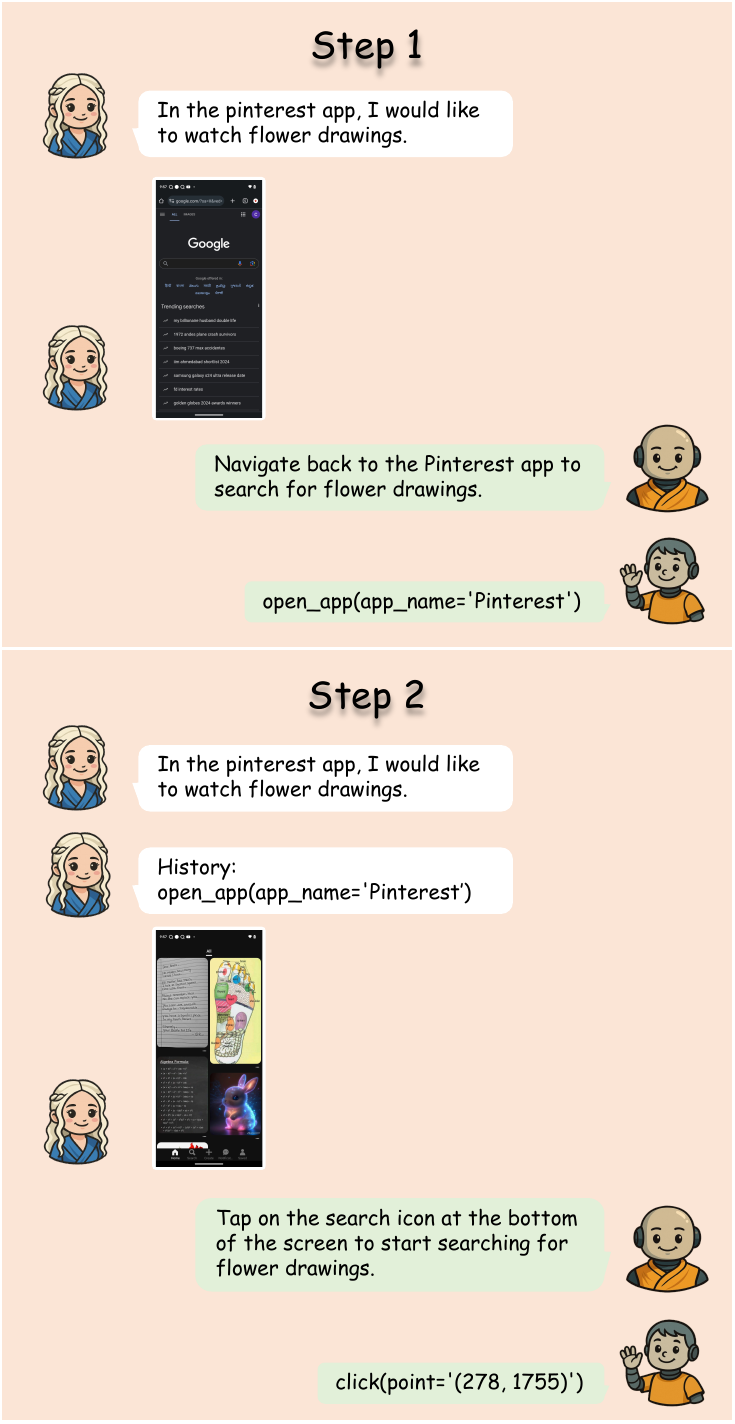}
    \caption{Example of GUI agents collaboratively performing a mobile GUI control task (Part 1 of 3). At each step, the system determines the next action based on the user instruction, the current screenshot, and the action history, iterating until the task is completed.}
    \label{fig:demo1}
\end{figure*}

\begin{figure*}[t!]
    \centering
    \includegraphics[width=0.65\textwidth]{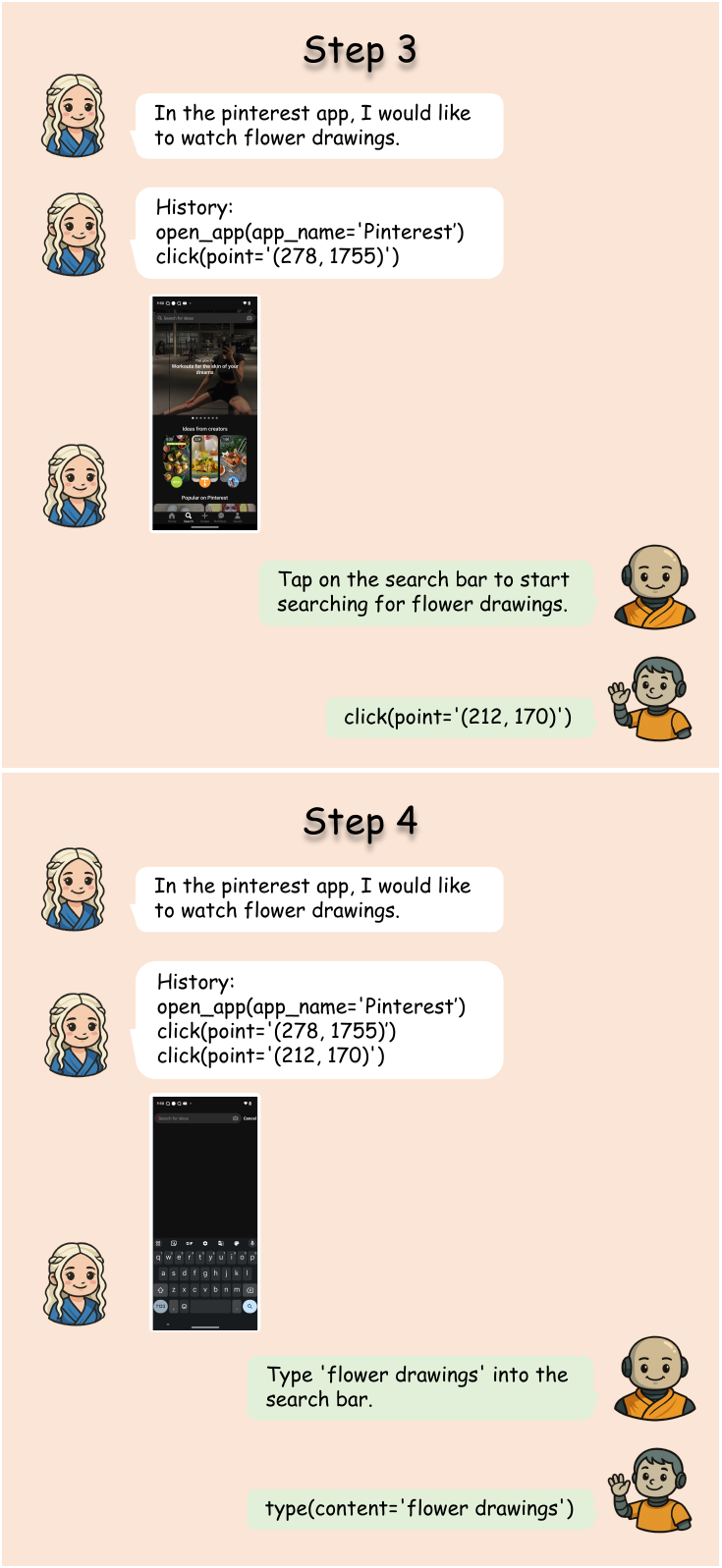}
    \caption{Example of GUI agents collaboratively performing a mobile GUI control task (Part 2 of 3).}
    \label{fig:demo2}
\end{figure*}

\begin{figure*}[t!]
    \centering
    \includegraphics[width=0.65\textwidth]{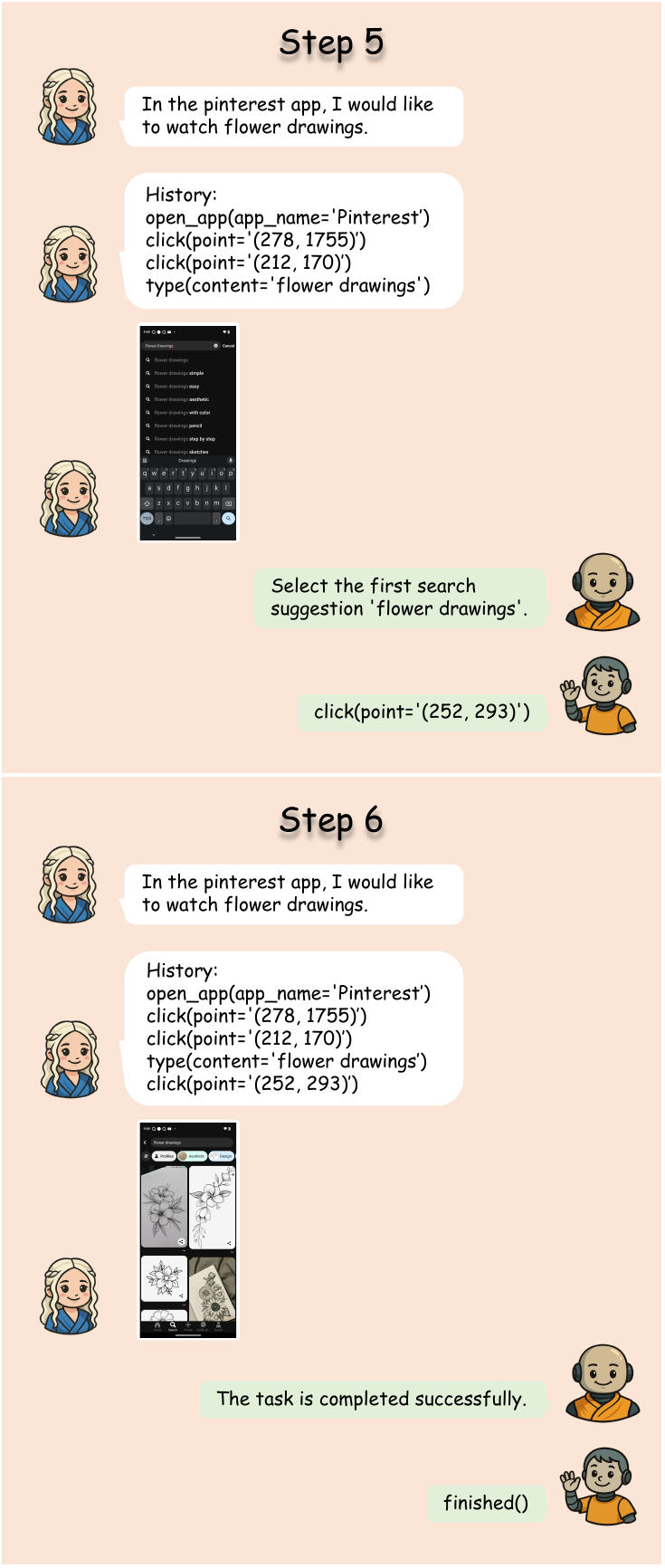}
    \caption{Example of GUI agents collaboratively performing a mobile GUI control task (Part 3 of 3).}
    \label{fig:demo3}
\end{figure*}

\end{document}